%% file: main.tex
\documentclass{article}
\PassOptionsToPackage{table,dvipsnames}{xcolor}
\usepackage{tikz}
\usetikzlibrary{calc,arrows.meta}

\PassOptionsToPackage{numbers,square,sort&compress}{natbib}
\usepackage[preprint]{neurips_2026}

\usepackage[utf8]{inputenc}
\usepackage[T1]{fontenc}
\usepackage{url}
\usepackage{booktabs}
\usepackage{amsfonts}
\usepackage{amssymb}
\usepackage{pifont}
\usepackage{graphicx}
\usepackage{adjustbox}
\usepackage{wrapfig}
\usepackage{makecell}
\usepackage{array}
\usepackage{enumitem}
\usepackage{nicefrac}
\usepackage{microtype}
\usepackage[table,dvipsnames]{xcolor}
\usepackage{amsmath}
\usepackage{multirow}
\usepackage{subcaption}
\usepackage{algorithm}
\usepackage{algorithmic}
\usepackage{arydshln}
\usepackage{placeins}
\usepackage{needspace}
\usepackage{accvabbrv}
\input{macros}

\usepackage{float}
\usepackage{tcolorbox}
\usepackage[breaklinks,colorlinks,citecolor=blue,linkcolor=blue,urlcolor=blue]{hyperref}
\hypersetup{pdftitle={Rubric-CEPR: Self-Evolving Image Editing via Reward-Verified Self-Distillation},pdfauthor={Ritesh Thawkar, Shubham Patle, Shravan Venkatraman, Rao Muhammad Anwer}}
\usepackage[nameinlink,capitalise,noabbrev]{cleveref}
\tcbuselibrary{skins,breakable}
\newtcolorbox{promptbox}{
  breakable,
  colback=blue!4,
  colframe=blue!45!black,
  boxrule=0.4pt,
  arc=2pt,
  left=5pt,
  right=5pt,
  top=4pt,
  bottom=4pt,
  before skip=2pt,
  after skip=2pt,
}

\definecolor{ncmetabg}{HTML}{F1F4F7}
\definecolor{ncmetaedge}{HTML}{DCE6F5}
\definecolor{ncmetablue}{HTML}{1877F2}
\definecolor{ncapricot}{HTML}{F7EBDD}

\newcommand{\githuburl}{https://github.com/riteshthawkar/Rubric-CEPR}
\newcommand{\projectpageurl}{https://riteshthawkar.github.io/Rubric-CEPR/}

\newcommand{\githubicon}{\raisebox{-0.18ex}{\makebox[1.35em][c]{\includegraphics[height=1.1em]{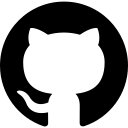}}}}
\newcommand{\projecticon}{\raisebox{-0.18ex}{\makebox[1.35em][c]{\includegraphics[height=1.1em]{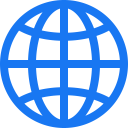}}}}

\newcommand{\xmark}{\ensuremath{\times}}

\newcount\mvcount
\newcommand{\mv}[1]{\global\advance\mvcount by 1\relax\textcolor{BrickRed}{#1}}
\AtEndDocument{\ifnum\mvcount>0 \GenericWarning{}{WARNING: manuscript contains MODELED red values (\string\mv)}\fi}

\newif\ifshowcolors
\showcolorsfalse
\ifshowcolors

\else

\fi

\makeatletter
\renewcommand{\@noticestring}{\noindent\rule{1.5in}{0.4pt}\\[2pt]
  \noindent$^{*}$Corresponding author: {\ttfamily ritesh.thawkar@mbzuai.ac.ae}}
\newcommand{\maketitleboxed}[1]{%
  \par
  \begingroup
    \renewcommand{\thefootnote}{\fnsymbol{footnote}}
    \renewcommand{\@makefnmark}{\hbox to \z@{$^{\@thefnmark}$\hss}}
    \long\def\@makefntext##1{%
      \parindent 1em\noindent
      \hbox to 1.8em{\hss $\m@th ^{\@thefnmark}$}##1%
    }
    \thispagestyle{empty}%
    \begin{tcolorbox}[
      enhanced,
      colback=ncmetabg,
      colframe=ncmetaedge,
      boxrule=0.35pt,
      arc=12pt,
      left=0.55cm, right=0.55cm, top=0.45cm, bottom=0.4cm,
      interior style={shade, shading angle=315,
        left color=white!96!ncmetabg,
        right color=ncmetablue!4!ncapricot!8!ncmetabg},
      before skip=0pt, after skip=0.4em,
      grow to left by=1.5pt, grow to right by=1.5pt,
    ]
      \centering
      {\LARGE\bf \@title\par}%
      \vskip 0.22in
      \def\And{\end{tabular}\hfil\linebreak[0]\hfil\begin{tabular}[t]{c}\bf\rule{\z@}{24\p@}\ignorespaces}%
      \def\AND{\end{tabular}\hfil\linebreak[4]\hfil\begin{tabular}[t]{c}\bf\rule{\z@}{24\p@}\ignorespaces}%
      \begin{tabular}[t]{c}\rule{\z@}{24\p@}\@author\end{tabular}\par
      \vskip 0.18in
      \begingroup
        \leftskip=1.5em \rightskip=1.5em
        \centerline{\large\bf Abstract}\vspace{0.6ex}
        \small #1\par
        \vspace{1.0ex}
        \begingroup
          \leftskip=1.5em \rightskip=1.5em plus 1fil \parfillskip=0pt \parindent=0pt
          \footnotesize
          \noindent\begin{tabular}{@{}c@{\hspace{0.55em}}l@{\hspace{0.35em}}l@{}}
            \githubicon & \textbf{Github Code:} & \href{\githuburl}{\texttt{\githuburl}}\\
            \projecticon & \textbf{Project Page:} & \href{\projectpageurl}{\texttt{\projectpageurl}}
          \end{tabular}\par
        \endgroup
      \endgroup
    \end{tcolorbox}%
    \@thanks
    \@notice
  \endgroup
  \let\maketitle\relax
  \let\thanks\relax
}
\makeatother

\title{Rubric-CEPR: Self-Evolving Image Editing via Reward-Verified Self-Distillation}

\author{%
Ritesh Thawkar$^{1,*}$ \quad
Shubham Patle$^{1}$ \quad
Shravan Venkatraman$^{1}$ \quad
Rao Muhammad Anwer$^{1,2}$\\[0.35em]
{\normalfont\small $^{1}$Mohamed bin Zayed University of Artificial Intelligence}\\
{\normalfont\small $^{2}$Aalto University \quad}
}

\input{figures/charts/style}
\input{appendix/setup}

\begin{document}
% Pages end flush: leftover space is spread over the (stretchable) vertical glue.
\flushbottom
\vbadness=10000 % page-stretch notices are expected with flush pages

\maketitleboxed{\input{sections/0_abstract}}
\enlargethispage{1\baselineskip}

% Keep the teaser below the abstract box and before the introduction.
\begin{figure}[H]
  \input{figures/teaser}
\end{figure}

\input{sections/1_introduction}
\input{sections/2_related_work}
\input{sections/3_method}
\input{sections/4_experiments}
\input{sections/5_results_and_analysis}

\input{sections/6_conclusion}

\section*{Acknowledgements}
The computations were enabled by resources provided by LUMI hosted by CSC (Finland) and LUMI consortium, and by Berzelius resource provided by the Knut and Alice Wallenberg Foundation at the NSC.

\bibliographystyle{plainnat}
\bibliography{main}

% Include the shared supplementary body after the complete reference list.
%
  \clearpage
  \raggedbottom
  \setlength{\textfloatsep}{10pt plus 2pt minus 2pt}%
  \setlength{\floatsep}{8pt plus 2pt minus 2pt}%
  \setlength{\intextsep}{8pt plus 2pt minus 2pt}%
  \appendix
  \setcounter{figure}{0}%
  \setcounter{table}{0}%
  \setcounter{equation}{0}%
  \renewcommand{\theHsection}{appendix.\Alph{section}}%
  \renewcommand{\theHsubsection}{\theHsection.\arabic{subsection}}%
  \renewcommand{\theHsubsubsection}{\theHsubsection.\arabic{subsubsection}}%
  \renewcommand{\thefigure}{A\arabic{figure}}%
  \renewcommand{\thetable}{A\arabic{table}}%
  \renewcommand{\theequation}{A\arabic{equation}}%
  \renewcommand{\thesuppalgorithm}{A\arabic{suppalgorithm}}%
  \renewcommand{\theHfigure}{appendix.\arabic{figure}}%
  \renewcommand{\theHtable}{appendix.\arabic{table}}%
  \renewcommand{\theHequation}{appendix.\arabic{equation}}%
  \renewcommand{\theHsuppalgorithm}{appendix.\arabic{suppalgorithm}}%
  \input{appendix/supplementary}
\end{document}

%% file: macros.tex
\usepackage{amsmath}
\usepackage{amssymb}
\usepackage{xspace}
\usepackage{dsfont}

\newcommand{\method}{Rubric-CEPR\xspace}
\newcommand{\qwenedit}{Qwen-Image-Edit\xspace}
\newcommand{\cepr}{CEPR\xspace}
\DeclareMathOperator{\gem}{gmean}
\newcommand{\indic}{\mathds{1}}

\definecolor{roleA}{RGB}{37,99,235}   % Proposer  (blue)
\definecolor{roleB}{RGB}{124,58,237}  % Editor    (violet)
\definecolor{roleC}{RGB}{16,133,72}   % Evaluator (green)
\definecolor{roleD}{RGB}{201,103,4}   % Update    (amber)
\definecolor{oursbg}{HTML}{EAF1FB}
\newcommand{\oursrow}{\rowcolor{oursbg}}

\newcommand{\tgroup}[2]{\multicolumn{#1}{c}{\emph{#2}}\\\hdashline\noalign{\vskip 3pt}}

\newcommand{\uprel}[1]{\textcolor{ForestGreen}{$+#1\%$}}

\newcommand{\relrow}{\quad\textit{$\Delta$ Improvement}}
\newcommand{\extrw}{$^{\dagger}$}

%% file: figures/charts/style.tex
\usepackage{pgfplots}
\pgfplotsset{compat=1.18}
\usetikzlibrary{patterns}
\definecolor{chartBlue}{HTML}{0057B8}
\definecolor{chartGreen}{HTML}{008543}
\definecolor{chartTeal}{HTML}{006B73}
\definecolor{chartOrange}{HTML}{C76300}
\definecolor{chartRed}{HTML}{D42B35}
\definecolor{chartPurple}{HTML}{7638B0}
\definecolor{chartInk}{HTML}{153E75}
\newcommand{\chartfont}{\fontsize{8}{9.5}\selectfont\sffamily}
\newcommand{\chartsmall}{\fontsize{7}{8.5}\selectfont\sffamily}
\newcommand{\charttitle}{\fontsize{9}{10.5}\selectfont\sffamily\bfseries}
\newcommand{\chartemph}{\chartfont\bfseries\boldmath}
\pgfplotsset{scientific chart/.style={
  font=\chartfont,
  axis lines*=left,
  axis line style={chartInk,line width=.5pt},
  tick style={chartInk,line width=.5pt},
  tick align=outside,
  ticklabel style={font=\chartfont},
  yticklabel style={align=right},
  label style={font=\chartfont},
  legend style={font=\chartfont,draw=none,fill=white,inner sep=2pt},
  legend cell align=left,
  ymajorgrids=false,
  grid style={chartInk,densely dotted,line width=.2pt},
  scaled ticks=false,
  ticklabel style={font=\chartfont,/pgf/number format/fixed},
  /pgf/number format/1000 sep={},
  line width=1pt,
  mark size=2pt,
  clip=false,
  enlargelimits=false,
  every axis plot/.append style={line join=round},
}}

\newlength{\chartmainaxiswidth}
\newlength{\chartmainaxisheight}
\newlength{\chartmainstackbarwidth}
\newlength{\chartmaintransferbarwidth}
\newcommand{\chartmainsetup}{%
  \setlength{\chartmainaxiswidth}{.225\linewidth}%
  \setlength{\chartmainaxisheight}{3.05cm}%
  \setlength{\chartmainstackbarwidth}{.215\chartmainaxiswidth}%
  \setlength{\chartmaintransferbarwidth}{.225\chartmainaxiswidth}%
  \renewcommand{\chartfont}{\fontsize{8}{9.5}\selectfont\sffamily}%
  \renewcommand{\chartsmall}{\fontsize{7}{8.5}\selectfont\sffamily}%
  \ifdim\linewidth<400pt
    \renewcommand{\chartfont}{\fontsize{7}{8.5}\selectfont\sffamily}%
    \renewcommand{\chartsmall}{\fontsize{6.5}{8}\selectfont\sffamily}%
    \renewcommand{\charttitle}{\fontsize{7.5}{9}\selectfont\sffamily\bfseries}%
  \fi
}
\pgfplotsset{matched main chart/.style={scientific chart,scale only axis,
  width=\chartmainaxiswidth,height=\chartmainaxisheight}}

%% file: appendix/setup.tex
\newcounter{suppalgorithm}
\renewcommand{\thesuppalgorithm}{S\arabic{suppalgorithm}}
\providecommand{\theHsuppalgorithm}{\arabic{suppalgorithm}}
\newcolumntype{L}[1]{>{\raggedright\arraybackslash}p{#1}}

\newenvironment{suppalgorithm}[1]{%
  \par\Needspace{4\baselineskip}\medskip\noindent
  \refstepcounter{suppalgorithm}%
  \hrule\vspace{3pt}
  \textbf{Algorithm~\thesuppalgorithm. #1}\par\vspace{2pt}\hrule\vspace{3pt}
  \begingroup\small
  \setlength{\abovedisplayskip}{2pt}%
  \setlength{\belowdisplayskip}{2pt}%
  \setlength{\abovedisplayshortskip}{1pt}%
  \setlength{\belowdisplayshortskip}{1pt}%
}{%
  \par\endgroup
  \vspace{2pt}\hrule\par\medskip
}

%% file: sections/0_abstract.tex
Instruction-guided image editors have become highly capable, yet improving them
further still depends on human-edited training pairs or external reward models.
Such supervision is costly to obtain and can reward plausible failures: a
realistic output may leave the requested change undone or alter content that
should be preserved. In this work, we strive to improve a pretrained image editor
using only its own generations, without human-edited targets or an external
training-time reward model. To this end, we propose a self-evolving framework,
named \method{}, that verifies the editor's own samples with its internal
representations through a rubric-augmented Contrastive Edit-Preservation Reward
(CEPR). A \emph{Planner} proposes structured edit instructions from unlabeled
images, the \emph{Editor} samples multiple candidate edits, and a frozen \emph{Critic} scores
each candidate with decomposed rubric checks for edit realization, removal of
the old state, and content preservation, using features already exposed by the
editor. Non-compensatory gates reject infeasible candidates, and the best
verified candidate is distilled into the editor through lightweight adapter
training. On Qwen-Image-Edit, \method{} improves ImgEdit from
$4.36$ to $4.60$ ($+5.5\%$), with a $+24.9\%$ gain on object isolation, and
transfers to GEdit-Bench and Complex-Edit. The same procedure also improves
Step1X-Edit by $+7.8\%$ on ImgEdit. We hope our approach will serve as a solid
baseline for image editors that improve themselves from their own verified
samples.

%% file: figures/teaser.tex
% Shared teaser content; placement is controlled by the manuscript root.
  \centering
  \input{figures/teaser_tikz}
  \caption{\textbf{Illustration of our self-evolving image editing framework (\method{}).}
Our \method{} improves a pretrained image editor \emph{without human-edited
targets or an external reward model}. Given only unlabeled images, a \emph{Planner}
proposes structured edit instructions, the \emph{Editor} samples $K$ candidate edits,
and an internal \emph{Critic} scores them with features already exposed by the editor.
Only the best gate-passing candidate per source becomes a target for weighted denoising
SFT, so the editor learns from its own verified successes. The resulting
single-shot checkpoint improves object isolation on ImgEdit from
$3.41\rightarrow4.26$ ($+24.9\%$) and the full ImgEdit score from
$4.36\rightarrow4.60$ ($+5.5\%$), and transfers to GEdit-Bench ($+12.4\%$).}
  \label{fig:teaser}

%% file: sections/1_introduction.tex
\section{Introduction}
\label{sec:introduction}

Instruction-guided image editing has progressed rapidly, from inversion- and
attention-based techniques~\cite{hertz2022prompt,meng2022sdedit,mokady2023nulltext}
to instruction-following editors built on multimodal
backbones~\cite{brooks2023instructpix2pix,sheynin2024emuedit,liu2025step1xedit,wu2025qwenimage,blackforestlabs2025fluxkontext}.
An editor must satisfy two requirements: perform the requested change and
preserve the content that the instruction leaves unchanged. Strong pretrained
editors still violate this contract. For example, an instruction to isolate a
shirt may produce a realistic photograph of the person wearing it, preserving
appearance while failing to isolate the object
(Figure~\ref{fig:qualitative_samples}). Improving editors beyond such failures
currently relies on two forms of external supervision: (i) curated pairs of
source and edited images~\cite{zhang2023magicbrush,zhao2024ultraedit,ye2025imgedit},
and (ii) learned reward models or multimodal judges that score generated
edits~\cite{wu2025editreward,luo2026editscore,li2025uniworldv2}. Both are costly
to scale, and a scalar measure of prompt relevance or visual plausibility can
still favor an unchanged source, a partial edit, or an attractive output that
discards important content.

Recent research in large language models explores self-evolving models that
generate their own training data and improve from their own
judgments~\cite{wang2023selfinstruct,zelikman2022star,yuan2024selfrewarding,chen2024spin}.
These methods show that a model can bootstrap from its own outputs when
correctness can be checked from text, through answer agreement, reasoning
traces, or a learned judge. In image editing, recent post-training methods
optimize editors with preference or reward
signals~\cite{wallace2024diffusiondpo,li2025uniworldv2,wu2025editreward,luo2026editscore,zhao2026imageeditr1}.
However, these approaches obtain their reward from a separately trained reward
model or an external multimodal judge. Self-training an editor is particularly
sensitive to this signal: plausible failures that pass a lenient score become
supervision for the next model.

In this work, we investigate the following question: \emph{Can a pretrained
image editor improve its own editing ability by verifying its generated edits
with its own internal representations, without human-edited targets or an
external reward model?} To this end, we propose \method{}, a self-evolving
framework that follows a propose--edit--verify--distill loop
(Figure~\ref{fig:teaser}) and scores candidates with a rubric-augmented
\emph{Contrastive Edit-Preservation Reward} (CEPR): the reward contrasts support
for the requested edit with incorrect alternative instructions and rewards
preservation of unrelated content, while the rubric adds explicit checks for
required states, removal of forbidden old states, and preservation constraints. A \emph{Planner} expresses the intended operation as an
instruction and a structured specification. The \emph{Editor} samples several
candidate edits, and a fixed \emph{Critic} checks the specification using the editor's
own multimodal features and latents. Edit realization, suppression of the old
state, preservation, and validity are checked separately, and failure of any
applicable check makes a candidate ineligible for training. The best verified
candidate then supervises a lightweight editor adapter through the editor's
native denoising objective. It is worth mentioning that the primary model never
queries a human-edited target or an external reward model during training.

We evaluate the resulting checkpoint in the standard single-shot setting. On
\texttt{Qwen-Image-Edit-2509}, \method{} improves ImgEdit from $4.36$ to $4.60$
($+5.5\%$), consistently across three training seeds, with a $+24.9\%$ gain on
object isolation, the weakest family of the base editor. The same checkpoint
transfers to GEdit-Bench and Complex-Edit, and applying the procedure to
Step1X-Edit improves ImgEdit by $+7.8\%$. Our analysis further shows that gains
are largest where the base editor is weak but its success criterion is
internally checkable.

\noindent In summary, our main contributions are:
\begin{itemize}[leftmargin=*,itemsep=2pt,topsep=3pt]
\item We introduce \method{}, a self-evolving image editing framework in which a
\emph{Planner}, an \emph{Editor}, and a fixed internal \emph{Critic} form a propose--edit--verify--distill
loop that improves a pretrained editor from unlabeled images, without
human-edited targets or an external training-time reward model.
\item We develop a decomposed rubric reward with non-compensatory feasibility
gates that checks edit realization, removal of the old state, and preservation
separately using editor-exposed Qwen-VL features and VAE latents, so that
plausible failures cannot become training targets.
\item We empirically validate \method{} on ImgEdit, GEdit-Bench, and
Complex-Edit, with a $+5.5\%$ ImgEdit gain and a $+24.9\%$ gain on object
isolation for \texttt{Qwen-Image-Edit-2509}, and a $+7.8\%$ ImgEdit gain on
Step1X-Edit. We further show that training the \emph{Planner} increases the rate of
feasible targets by $62.4\%$.
\end{itemize}

%% file: sections/2_related_work.tex
\section{Related Work}
\label{sec:related_work}

\paragraph{Instruction-guided image editing.}
Inversion, attention control, and feature manipulation enable edits to real
images at inference time~\cite{meng2022sdedit,avrahami2022blended,couairon2023diffedit,hertz2022prompt,mokady2023nulltext,tumanyan2023plugandplay,kawar2023imagic,parmar2023pix2pixzero,cao2023masactrl}.
Instruction-following editors extend this interface to natural-language commands
and multimodal backbones~\cite{brooks2023instructpix2pix,geng2024instructdiffusion,fu2024mgie,huang2024smartedit,sheynin2024emuedit,zhao2024ultraedit,liu2025step1xedit,ye2025imgedit,wu2025qwenimage,blackforestlabs2025fluxkontext},
and are commonly evaluated on ImgEdit, GEdit-Bench, and
Complex-Edit~\cite{ye2025imgedit,liu2025step1xedit,yang2025complexedit}.
These editors are trained on large collections of curated source--target
pairs~\cite{brooks2023instructpix2pix,zhang2023magicbrush,zhao2024ultraedit,ye2025imgedit}.
However, once pretrained, they provide no mechanism to keep improving from
their own outputs without collecting new paired data.

\paragraph{Reward models and editing post-training.}
Human and text--image preference models provide supervision for image
generation~\cite{xu2024imagereward,kirstain2023pickapic,wu2023hpsv2,lee2023aligning}.
EditReward~\cite{wu2025editreward} trains an editing-specific reward model on
expert preference annotations and uses it to filter training data for Step1X.
EditScore~\cite{luo2026editscore} connects editing-reward benchmarking,
best-of-$N$ selection, and online reinforcement learning, and UniWorld's Edit-R1
framework~\cite{li2025uniworldv2} optimizes an editor with the logits of an
external multimodal LLM through diffusion negative-aware fine-tuning.
ImageEdit-R1 and NP-Edit also study reward-based editing
updates~\cite{zhao2026imageeditr1,kumari2025npedit}, while UniEdit-I couples
understanding, editing, and verification at inference time~\cite{bai2025uniediti}.
While effective, these approaches require either a separately trained reward
model, annotated preferences, or an external multimodal judge, and a single
scalar score can still reward plausible outputs that do not realize the
requested edit.

\paragraph{Self-training and preference optimization.}
Iterative self-generation and self-rewarding improve language models through
instruction generation, reasoning, or preference
training~\cite{wang2023selfinstruct,zelikman2022star,yuan2024selfrewarding,chen2024spin}.
Self-Rewarding Language Models~\cite{yuan2024selfrewarding} update both response
generation and judging ability within the same model. In multimodal models,
self-rewarding and self-evolving frameworks let the understanding branch
supervise generation~\cite{jin2026srum,hong2025suder,han2026unicorn,thawkar2026asg},
and CVPD distills a model's own counterfactual visual evidence into dense
perception supervision~\cite{venkatraman2026cvpd}. For editing, JarvisEvo
co-optimizes a photo-editing agent as editor and evaluator~\cite{lin2025jarvisevo},
and MT-OPSD applies on-policy self-distillation to multi-turn
editing~\cite{zhao2026mtopsd}. DPO and its diffusion
variants learn from preference
pairs~\cite{rafailov2023dpo,wallace2024diffusiondpo,yang2024d3po}, and RL or
differentiable-reward methods optimize rewards through
denoising~\cite{fan2023dpok,black2024ddpo,clark2024draft}. Verifier-guided
search~\cite{ma2025inferencetimescaling} improves samples at inference time,
and reward over-optimization~\cite{gao2023overoptimization} shows that policies
can exploit imperfect rewards. However, these methods target language,
perception, text-to-image generation, tool-based retouching, or multi-turn
consistency, and none verifies single-turn instruction edits against a structured
specification, where the reward must distinguish a correct edit from a plausible
but unedited output.

\noindent\textbf{Our Approach.}
Different from the aforementioned approaches, \method{} improves a pretrained
image editor \emph{without} human-edited targets or an external training-time
reward model. Our framework uses only unlabeled source images and a
\emph{Planner}--\emph{Editor}--\emph{Critic} loop where (i) the \emph{Planner} proposes structured edit
specifications, (ii) the \emph{Editor} samples candidate edits, and (iii) a fixed
\emph{Critic} verifies each candidate with the editor's own Qwen-VL features, VAE
latents, and simple image statistics. Instead of a single scalar score, the
\emph{Critic} checks edit realization, removal of the old state, and preservation
separately, and non-compensatory gates reject any candidate that fails an
applicable check. Unlike self-rewarding methods whose judge changes with the
policy, our \emph{Critic} stays fixed, so the selection criterion does not drift as the
editor adapts. Verified targets are distilled with the editor's native
denoising objective, and the adapted editor is evaluated in the standard
single-shot setting. Published results of reward-based editing methods are
reported with their own base and protocol in
Tables~\ref{tab:main_results} and~\ref{tab:transfer_benchmark_results}.

%% file: sections/3_method.tex
\section{Method}
\label{sec:method}

\paragraph{Problem Formulation.}
We consider the setting of \emph{self-evolving} image editing, where no
human-edited targets or external reward models are available during training.
Let $\mathcal X=\{x\}$ denote a collection of unlabeled source images and $E_0$
a pretrained instruction-guided editor. Our goal is to improve $E_0$ using
training targets constructed from its own verified samples. To this end, we
instantiate three components:
\begin{itemize}[leftmargin=*,itemsep=1pt,topsep=2pt]
\item a \textbf{\emph{Planner}} $P_\psi(c,s\mid x)$ that proposes a natural-language
instruction $c$ and a structured edit specification $s$ for image $x$;
\item an \textbf{\emph{Editor}} $E_\theta(y\mid x,c)$ that samples candidate edits $y$;
and
\item a fixed \textbf{\emph{Critic}} $C(x,c,s,y)$ that returns a reward $R$ and a
feasibility indicator $G\in\{0,1\}$ from the editor's own features.
\end{itemize}
The \emph{Planner} and \emph{Editor} are trained through lightweight LoRA adapters, while the
reference feature interface and the \emph{Critic} remain frozen.

\begin{figure*}[t]
  \centering
  % Overlay corrects two labels in the vector figure: "Best of N" -> "Best-of-K", "Round (R_{t+1})" -> "Round (t+1)".
  \begin{tikzpicture}[x=\linewidth,y=\linewidth,inner sep=0pt]
    \node[anchor=south west] at (0,0) {\includegraphics[width=\linewidth]{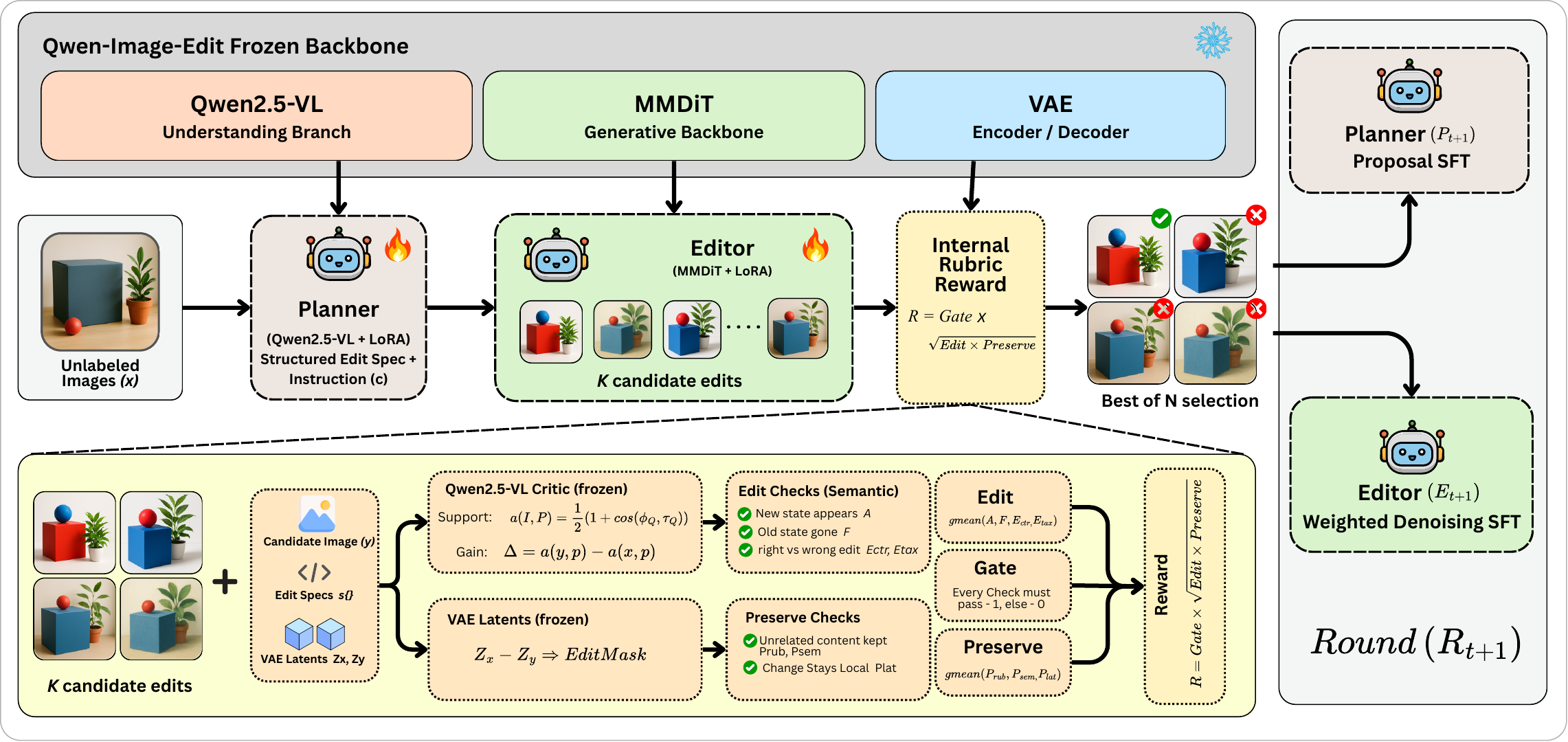}};
    \fill[white] (0.699,0.2135) rectangle (0.806,0.2245);
    \node at (0.7525,0.2188) {\sffamily\bfseries\fontsize{5.8pt}{6pt}\selectfont Best-of-$K$ selection};
    \fill[fill={rgb,255:red,244;green,246;blue,245}] (0.832,0.0500) rectangle (0.972,0.0750);
    \node at (0.9027,0.0625) {\fontsize{9pt}{10pt}\selectfont $\mathit{Round}\,(t{+}1)$};
  \end{tikzpicture}
  \caption{\textbf{Overview of the \method{} propose--edit--verify--distill loop.}
The frozen \qwenedit{} backbone provides the Qwen2.5-VL understanding branch,
the MMDiT generative backbone, and the VAE. Given an unlabeled image $x$, the
\emph{Planner} (Qwen2.5-VL with LoRA) emits an instruction $c$ and a structured edit
specification $s$, and the \emph{Editor} (MMDiT with LoRA) samples $K$ candidate edits.
The frozen \emph{Critic} converts $s$ into rubric prompts and scores each candidate
with Qwen-VL support and gain (Eq.~\eqref{eq:support}) and VAE-latent
preservation checks. Semantic edit checks (the new state appears, the old state
is gone, the right edit is made) and preservation checks (unrelated content is
kept, the change stays local) form the non-compensatory reward
$R=\mathrm{Gate}\cdot\sqrt{\mathrm{Edit}\times\mathrm{Preserve}}$, where a
candidate is rejected if any applicable check fails. The best gate-passing
candidate trains the \emph{Editor} with weighted denoising SFT, and band-pass proposal
traces train the \emph{Planner}, giving $P_{t+1}$ and $E_{t+1}$ for the next round.
No external reward model is queried at any stage of the primary loop.}
  \label{fig:architecture}
\end{figure*}

\paragraph{Motivation.}
A natural baseline is to fine-tune the editor on its own generated candidates,
either directly or ranked by a single scalar score. We observe that this fails
for image editing: plausible but incorrect edits are common, and a scalar score
cannot tell an unchanged or partially edited image from a correct one. A naive
four-round self-training loop therefore improves the base editor less than verified targets do,
as rejected candidates outnumber usable targets (Section~\ref{subsec:target_selection}).
In \method{}, we instead verify every candidate against the structured
specification with decomposed checks and non-compensatory gates, and distill
only verified targets into the editor, as shown in
Figure~\ref{fig:architecture}. The primary instantiation targets object
isolation, the weakest family of the base editor; broader verifier-bank variants
are analyzed separately.

\subsection{Structured Proposals and Candidate Selection}
\label{sec:rounds}
\label{sec:proposals}
For a source $x$, the \emph{Planner} produces an instruction $c$ and specification
$s$. The instruction is the natural-language command given to the \emph{Editor}.
The specification adds machine-readable fields: edit type, source and target
entities, target region, required and forbidden after-states, and content to
preserve. For example, \emph{isolate the red shirt} specifies a complete shirt
as the required output, the person and surrounding scene as forbidden output
content, and shirt appearance as a preservation constraint. These fields
determine the \emph{Critic}'s atomic prompts; they are not a second edited target.

At round $t$, the current editor samples $K$ candidates
$y_i\sim E_t(\cdot\mid x,c)$ and the \emph{Critic} returns reward $R_i$ and feasibility
$G_i$. We retain one target:
\begin{equation}
\label{eq:selection}
 y^\star=\arg\max_{y_i:G_i=1} R_i,\qquad
 \mathcal D_t^+=\{(x,c,y^\star,R^\star):\exists i,\ G_i=1\}.
\end{equation}
If no candidate passes, that proposal contributes no editor target.
The primary self-distillation run uses $K=4$.
The general loop may repeat proposal generation, sampling, and adapter updates;
we evaluate the resulting checkpoint with one generation per instruction.
Algorithms~\ref{alg:supp_loop}--\ref{alg:supp_reward} give the complete loop.

\subsection{Frozen Internal Features}
\label{sec:features}
The image-conditioned feature $\phi_Q(I,p)$ and text anchor $\tau_Q(p)$ are
mean-pooled final-layer states of the frozen reference Qwen-VL component,
obtained with fixed multimodal and text-only templates. We compute
\begin{equation}
\label{eq:support}
 a(I,p)=\tfrac12\left[1+\cos(\phi_Q(I,p),\tau_Q(p))\right],
 \qquad d_p(y,x)=a(y,p)-a(x,p).
\end{equation}
Both features have the same dimensional representation, so no cross-model
projection is required. This permits a similarity calculation; it does not
guarantee calibrated edit correctness, which we examine with selection probes.
The reference VAE encodes $z_x$ and $z_y$ for latent preservation and locality.

For the primary object-isolation reward, these features are supplemented by
programmatic background-purity and object-completeness checks. No separate
learned reward model or external VLM is queried for its targets.
The broader verifier bank additionally uses yes/no token log-probabilities
from the editor-side VLM. Its addition analysis uses an open-vocabulary
detector, so that analysis variant is distinct from the primary internal-only
instantiation.

\subsection{Decomposed Rubric and Feasibility Gates}
\label{sec:rubric_reward}
Fixed templates convert $s$ into prompt sets for source grounding $S$,
required after-state $A$, forbidden after-state $F$, and explicit preservation
$P_{\mathrm{rub}}$. Grounding checks the source entity before the edit.
Required-state prompts measure support and gain for the requested outcome.
Forbidden-state prompts measure both loss of old-state support and its absolute
absence. Preservation prompts compare the named unchanged content before and
after the edit. Applicable terms aggregate shifted-sigmoid supports
geometrically; inapplicable terms are omitted.

Two additional edit terms compare the true instruction with distractors:
$E_{\mathrm{ctr}}$ checks gain against counterfactual instructions, and
$E_{\mathrm{tax}}$ compares required, old-state, and wrong-edit prompts.
The scalar preservation term combines semantic and latent preservation,
$P_{\cepr}=\gem(P_{\mathrm{sem}},P_{\mathrm{lat}})$. Together with $E_{\mathrm{ctr}}$,
this forms the Contrastive Edit-Preservation Reward (CEPR), which compares support
for the requested edit against incorrect alternative instructions and rewards
preserved content; the rubric prompts above extend it with explicit required-state,
forbidden-state, and preservation checks.
For applicable components,
\begin{equation}
\label{eq:rubric}
 E=\gem(A,F,E_{\mathrm{ctr}},E_{\mathrm{tax}}),\qquad
 P=\gem(P_{\cepr},P_{\mathrm{rub}}),\qquad R=G\sqrt{EP}.
\end{equation}
Here $\gem$ is the geometric mean. A geometric score alone still permits
partial compensation; feasibility therefore requires each applicable check:
\begin{equation}
\label{eq:gates}
 G=\left(\prod_{u\in\mathcal U}\indic[u\geq\tau_u]\right)
   \indic[\sqrt{EP}\geq\tau_R],
 \quad
 \mathcal U\subseteq
 \{S,A,F,P_{\mathrm{rub}},E_{\mathrm{ctr}},E_{\mathrm{tax}},P_{\cepr},V,J\}.
\end{equation}
$J$ is included only for supported verifier-bank checks. A failed gate makes
the candidate ineligible for Eq.~\eqref{eq:selection}; visual plausibility cannot
override a failed edit or preservation check.

The validity term $V$ uses a relative latent-change mask
$M=\indic[|z_x-z_y|_{\mathrm{norm}}>\delta]$. This mask estimates changed
regions rather than supplying a ground-truth edit region. Sub-threshold
differences remain outside $M$; the check penalizes their magnitude, mismatch
with the area prior, and global drift. Thresholds and implementation settings
are provided in Appendix~\ref{sec:supp_implementation}.

\subsection{\texorpdfstring{\emph{Editor}}{Editor} and \texorpdfstring{\emph{Planner}}{Planner} Updates}
\label{sec:editor_objective}
For an accepted record, let $\ell_{\mathrm{FM}}^{\mathrm{native}}$ denote the
official backend's flow-matching denoising loss toward the selected target:
\begin{equation}
 \ell_{\mathrm{FM}}^{\mathrm{native}}(\theta;x,c,y)
 =\mathbb E_{t,\epsilon}\!\left[
 \eta(t)\|v_\theta(z_t,t;z_x,c)-u_t\|_2^2\right].
\end{equation}
The noisy target latent $z_t$, flow target $u_t$, and time weighting $\eta(t)$
follow the backend's scheduler and loss convention. Writing $\omega_r$ for a
record's training weight, the weighted adapter update has the form
\begin{equation}
\label{eq:editor_loss}
 \mathcal L_E=
 \frac{\sum_{r\in\mathcal D_t^+\cup\mathcal D_t^{\mathrm{id}}}
 \omega_r\,\ell_{\mathrm{FM}}^{\mathrm{native}}(\theta;x_r,c_r,y_r)}
 {\sum_r\omega_r}
 +\lambda_{\mathrm{anc}}\|A_\theta-A_{\mathrm{warm}}\|_F^2.
\end{equation}
$\mathcal D_t^{\mathrm{id}}$ contains identity/reconstruction replay pairs.
$A_\theta$ is the attention-only rank-16, $\alpha=16$ LoRA adapter, anchored to
the warm-start adapter $A_{\mathrm{warm}}$ with a relative-change cap.
Targets and rewards are detached; rejected candidates do not enter the loss.
The object-isolation run uses 400 SFT steps and learning rate $10^{-4}$.

The \emph{Planner} learns normalized instruction/specification JSON with
reward-weighted token SFT:
\begin{equation}
\label{eq:planner_loss}
 \mathcal L_P=-\frac{1}{Z_P}\sum_{j\in\mathcal B_t}\beta_j
       \log P_\psi(c_j,s_j\mid x_j),\qquad Z_P=\sum_{j\in\mathcal B_t}\beta_j .
\end{equation}
The band-pass set $\mathcal B_t$ excludes proposals for which every candidate
fails or every candidate passes, and retains traces satisfying the proposal
reward and quality thresholds. The \emph{Planner} adapter has rank 16, $\alpha=32$,
learning rate $10^{-5}$, and 16 training steps. This update improves proposal
feasibility in the matched control of Section~\ref{subsec:planner_control}.

%% file: sections/4_experiments.tex
\section{Experimental Setup}
\label{sec:experiments}

\subsection{Training Data and Implementation Details}
Our training uses only unlabeled, off-benchmark source images; no benchmark
image or human-edited target is used for supervision. All editor targets are
the editor's own candidates selected by the \emph{Critic}. The primary base is
\texttt{Qwen-Image-Edit-2509}. The \emph{Editor} is adapted with an attention-only
rank-16 LoRA ($\alpha=16$) for 400 SFT steps with learning rate $10^{-4}$, using
$K=4$ candidates per proposal; the \emph{Planner} uses a rank-16 LoRA ($\alpha=32$)
trained for 16 steps with learning rate $10^{-5}$. A second instantiation applies
the same procedure to Step1X-Edit with its own VLM and VAE, without importing
\qwenedit{} weights. Table~\ref{tab:supp_impl_settings} lists all generation,
adapter, and verifier settings.

\subsection{Evaluation}
We evaluate on the full ImgEdit, GEdit-Bench, and Complex-Edit benchmarks with
one generation per instruction and matched generation settings for the base and
adapted editors. ImgEdit is scored with GPT-4o ratings on the 0--5 scale
(Table~\ref{tab:main_results}). GEdit-Bench uses all 11 tasks with
mixed-language prompts scored by GPT-4.1 VIEScore, and Complex-Edit uses its
compound edits and 0--10 metrics (Table~\ref{tab:transfer_benchmark_results}).
For Step1X-Edit, we report GEdit-Bench and ImgEdit
(Table~\ref{tab:step1x_transfer}). Results taken
from other papers keep their original protocols and are not rebased onto our
scores. The GEdit-Bench background task was also used when choosing reward types.

\input{tables/main_results}

\subsection{Baselines and Controls}
We compare against the pretrained base editor evaluated under the same protocol,
published image editors, and UniWorld~\cite{li2025uniworldv2}, which post-trains
the same base with an external multimodal reward. To analyze the components of
\method{}, we further use: (i) a naive four-round self-training loop with two
candidates per proposal and no feasibility gates; (ii) a fixed 210-pair probe
that tests candidate ranking and negative-control acceptance; and (iii) a \emph{Planner}
control that changes only whether the \emph{Planner} is trained, together with a
blinded 64-source audit of proposal grounding. We report the sample standard
deviation (SD) over three training seeds where multiple seeds were trained, and
otherwise the bootstrap standard error (s.e.) of the paired score difference over
test items (Step1X-Edit ImgEdit).

%% file: tables/main_results.tex
\begin{table}[!t]
\centering
\caption{\textbf{Comparison with image editing methods on ImgEdit.}
We compare published editors, UniWorld (external reward), and our \method{},
which self-distills \qwenedit{} with only internal rewards. \method{} improves
every family and raises the overall score from $4.36\rightarrow4.60$ ($+5.5\%$;
$+0.24\pm0.02$ SD over three seeds), with the largest gain on extraction
($3.41\rightarrow4.26$, $+24.9\%$), the weakest base family. Scores are on the $0$--$5$ scale and $\Delta$ Improvement
is the relative improvement over the base,
$100\times(\text{adapted}-\text{base})/\text{base}$. \extrw{}UniWorld post-trains
\qwenedit{} with an external multimodal-LLM reward model (training-free, logit-based)
and reports $4.35\rightarrow4.48$ ($+3.0\%$) over its own base. Rows from other papers use
their own judges and are not directly comparable.}
\label{tab:main_results}
\small
\setlength{\tabcolsep}{2.85pt}
\begin{tabular}{@{}lcccccccccc@{}}
\toprule
\textbf{Model} & \textbf{Add} & \textbf{Remove} & \textbf{Replace} &
\textbf{Adjust} & \textbf{Action} & \textbf{Style} & \textbf{Bg.} &
\textbf{Extract} & \textbf{Compose} & \textbf{Overall} \\
\midrule
\tgroup{11}{Image editing models}
GPT-4o-Image~\cite{ye2025imgedit} & \textbf{4.65} & 3.81 & 4.49 & 4.26 & 4.76 & 4.75 & \textbf{4.62} & 2.96 & \textbf{4.54} & 4.32 \\
Step1X-Edit~\cite{liu2025step1xedit} & 3.90 & 2.61 & 3.45 & 3.13 & 3.43 & 4.44 & 3.19 & 1.87 & 2.52 & 3.17 \\
ImgEdit-E1~\cite{ye2025imgedit} & 3.82 & 2.40 & 2.80 & 4.04 & 3.21 & 4.38 & 3.38 & 2.55 & 2.87 & 3.27 \\
UltraEdit~\cite{zhao2024ultraedit} & 3.63 & 1.71 & 3.13 & 3.01 & 3.57 & 3.69 & 3.31 & 2.02 & 2.33 & 2.93 \\
MagicBrush~\cite{zhang2023magicbrush} & -- & -- & -- & -- & -- & -- & -- & -- & -- & 1.90 \\
InstructPix2Pix~\cite{brooks2023instructpix2pix} & 2.29 & 1.49 & 1.93 & 1.79 & 1.51 & 3.54 & 1.67 & 1.33 & 1.48 & 1.89 \\
\midrule
\tgroup{11}{Reward post-trained methods}
UniWorld-Qwen-Edit\extrw~\cite{li2025uniworldv2} & 4.34 & 4.65 & 4.62 & 4.40 & 4.80 & 4.89 & 4.31 & 4.37 & 3.90 & 4.48 \\
UniWorld-V2\extrw~\cite{li2025uniworldv2} & 4.29 & 4.72 & 4.69 & 4.44 & 4.83 & 4.91 & 4.41 & 4.32 & 3.83 & 4.49 \\
\midrule
Qwen-Image-Edit-2509~\cite{wu2025qwenimage} & 4.51 & 4.36 & 4.72 & 4.40 & 4.69 & 4.70 & 4.40 & 3.41 & 4.05 & 4.36 \\
\oursrow \method{} (ours) & \textbf{4.65} & \textbf{4.50} & \textbf{4.86} & \textbf{4.54} & \textbf{4.82} & \textbf{4.83} & 4.59 & \textbf{4.26} & 4.39 & \textbf{4.60} \\
\oursrow \relrow & \uprel{3.1} & \uprel{3.2} & \uprel{3.0} & \uprel{3.2} & \uprel{2.8} & \uprel{2.8} & \uprel{4.3} & \uprel{24.9} & \uprel{8.4} & \uprel{5.5} \\
\bottomrule
\end{tabular}
\end{table}

%% file: sections/5_results_and_analysis.tex
\section{Results and Analysis}
\label{sec:results}

We first report the main results on ImgEdit, the transfer to GEdit-Bench,
Complex-Edit and a second editor, then analyze why verified targets matter and
what training the \emph{Planner} adds.

\subsection{Main Results}
\label{subsec:posttraining}
\label{subsec:self_distill}
In Table~\ref{tab:main_results}, we compare \method{} with three groups of
baselines: (1) the pretrained \qwenedit{} base without additional training,
evaluated under our protocol; (2) published image editors; and (3) UniWorld,
which post-trains the same base with an external multimodal reward.

\noindent\textbf{\qwenedit{} (Base).}
The base editor is already strong on many edit types, such as action (4.69),
replacement (4.72), and style (4.70).
However, its performance drops on edits that require removing surrounding
content while keeping the target intact, most notably extraction (3.41) and
composition (4.05), indicating that pretraining alone does not reliably enforce
the edit contract.

\noindent\textbf{\method{} (Ours).}
Self-distilling verified targets improves every edit family and raises the
overall ImgEdit score from 4.36 to 4.60 ($+5.5\%$). The adapted score is stable
across three training seeds (4.58, 4.60, and 4.62; $4.60\pm0.02$ SD), giving a
gain of $+0.24\pm0.02$ SD. The
largest improvement is on extraction, which rises from 3.41 to 4.26
($+24.9\%$), followed by composition ($+8.4\%$); the remaining seven families
gain between $+2.8\%$ and $+4.3\%$. Measured against the headroom left below the
$5.0$ ceiling, \method{} closes between $22\%$ and $53\%$ of the remaining gap on every
family (Figure~\ref{fig:ablation_summary}c), so the large extraction gain largely
reflects its large headroom. UniWorld reports $+3.0\%$ over its own base
with an external multimodal reward; \method{} obtains a larger relative gain
using only internal signals, although the two evaluation protocols differ.

\noindent\textbf{Transfer to other benchmarks.}
The same checkpoint transfers without further adaptation
(Table~\ref{tab:transfer_benchmark_results}). GEdit-Bench improves from 7.39 to
8.31 ($+12.4\%$) and Complex-Edit from 8.77 to 8.97 ($+2.3\%$), and every reported
submetric improves, including identity preservation ($+1.8\%$) and perceptual
quality ($+4.9\%$) on Complex-Edit. At the task level, all eleven GEdit-Bench tasks improve, with the largest gains
on style change, human photo editing, and tone transfer
(Table~\ref{tab:supp_gedit_tasks}).

\input{tables/transfer_benchmark_results}

\noindent\textbf{Generalization to a second editor.}
\label{subsec:step1x_transfer}
Applying the same procedure to Step1X-Edit improves the overall GEdit-Bench
score from 6.69 to 7.24 ($+8.2\%$; $+0.55\pm0.05$ SD across three training seeds)
and ImgEdit from 3.86 to 4.16 ($+7.8\%$; $+0.30\pm0.06$ s.e.), improving the
overall score on both benchmarks (Table~\ref{tab:step1x_transfer}). Since Step1X-Edit
uses its own VLM and VAE, this shows that internal verification is not specific
to a single editor.

\input{tables/step1x_transfer}

\subsection{Why Verified Targets Matter}
\label{subsec:target_selection}

\noindent\textbf{Naive self-training improves less.}
Training on the editor's own candidates without feasibility gates yields a
smaller ImgEdit gain ($+0.09$, $+2.0\%$) than \method{} ($+5.5\%$), and a smaller
GEdit-Bench diagnostic gain ($+0.77$, $+10.5\%$ vs.\ $+12.4\%$). The training manifest (Appendix~\ref{sec:supp_self_evolution_analysis},
Figure~\ref{fig:diagnostic_evidence}) explains this behavior: rejected candidates outnumber
accepted ones, and the few clean targets are dominated by easy color edits,
while removal and replacement receive almost none.

\begin{figure*}[!htbp]
\centering
\input{figures/charts/ablation_summary}
\caption{\textbf{Verified targets, recoverable headroom, and where the gains land.}
(a) \emph{Valid targets.} On a fixed $210$-pair probe, a scalar reward still
assigns $0.39$--$0.60$ to no-op, corrupted, and wrong-edit candidates, whereas our
feasibility gates accept $51\%$ of valid edits and none of the invalid ones.
(b) \emph{Recoverable headroom.} On the extraction (object-isolation) family,
reward-best selection from $K$ base samples raises the score from $3.41$
($K{=}1$) to $3.89$, $4.22$, and $4.45$ ($K{=}2,4,8$), and the single-shot
distilled checkpoint reaches $4.26$ ($+24.9\%$), on par with best-of-4 without any
inference-time sampling. (c) \emph{Headroom closed.} Share of the gap between the base score and
the $5.0$ judge ceiling that \method{} closes on each ImgEdit family,
$(\text{ours}-\text{base})/(5-\text{base})$ from Table~\ref{tab:main_results}.
Extraction's large relative gain ($+24.9\%$) largely reflects its large headroom:
it closes $53\%$ of the gap, similar to replacement ($50\%$) and style ($43\%$),
and every family closes between $22\%$ and $53\%$, a $2.4\times$ spread compared
with $9\times$ for the raw relative gains.}
\label{fig:ablation_summary}
\end{figure*}
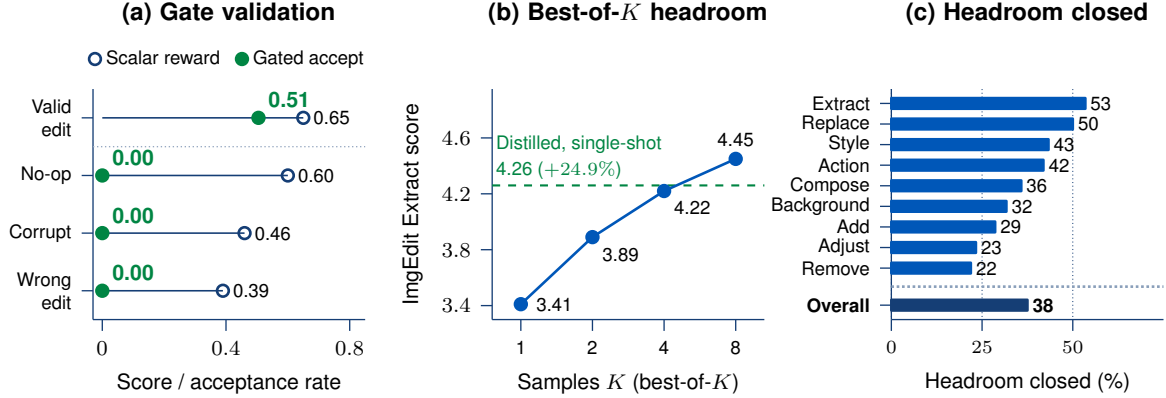

\noindent\textbf{Gates keep plausible failures out of training.}
A useful selector must both rank valid edits and reject invalid ones. On the
fixed 210-pair probe, our reward chooses the stronger edit in $84.8\%$ of pairs,
compared with $36.7\%$ for a naive product of its components
(Table~\ref{tab:supp_target_selection}). The full verifier-bank gate accepts
$50.5\%$ of valid edits and none of the no-op, corrupted, or wrong-edit controls,
whereas a scalar reward still assigns them scores of 0.39--0.60
(Figure~\ref{fig:ablation_summary}a). Removing the editor-side VLM check admits
$4.8\%$ of corrupted candidates (Table~\ref{tab:reward_gate_ablation}).

\input{tables/target_selection}

\begin{figure*}[!t]
\centering
\includegraphics[width=\textwidth]{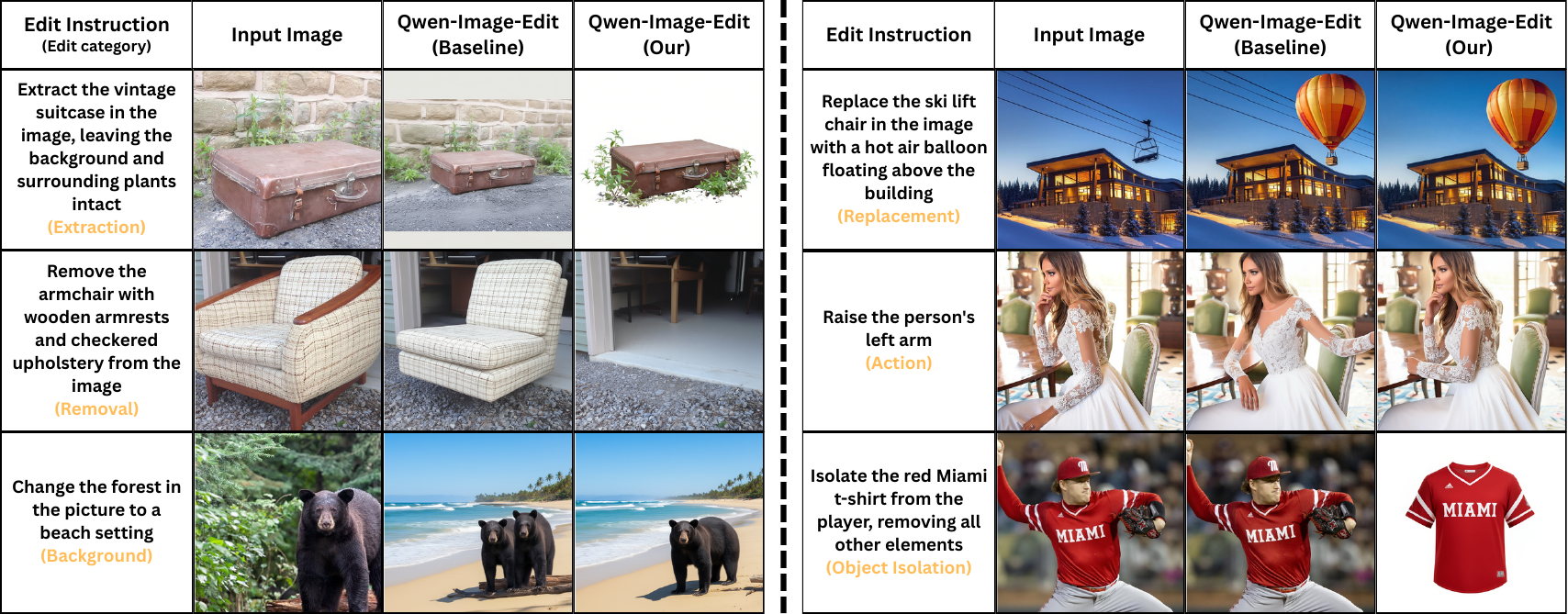}
\caption{\textbf{Qualitative comparison between \qwenedit{} and our \method{} on
ImgEdit.} Each row shows the edit instruction with its category, the input
image, the baseline output, and ours. The baseline often returns a realistic
image that leaves the requested change undone; for example, it keeps the player
when asked to isolate the red shirt (bottom right). \method{} completes such
edits while preserving the content that the instruction leaves unchanged.
Additional examples are provided in Appendix~\ref{sec:supp_qualitative}.}
\label{fig:qualitative_samples}
\end{figure*}

\noindent\textbf{Self-distillation internalizes best-of-$K$ headroom.}
For the primary isolation reward, mean border whiteness is 0.208 for outputs
that the evaluation judge scores at least 4, compared with 0.014 for outputs
scored at most 2, showing that the internal check tracks edit success.
Reward-best selection from $K{=}1,2,4,8$ base samples raises the extraction
score from 3.41 to 3.89, 4.22, and 4.45, and training on verified four-candidate
targets reaches 4.26 with a single generation, on par with best-of-4
(Figure~\ref{fig:ablation_summary}b). On the full extraction
family the matched change is 3.41 to 4.26 ($+0.85$, $+24.9\%$), with
71/27/19 wins/ties/losses against the base. Figure~\ref{fig:qualitative_samples} shows
representative edits.

\subsection{Proposal Learning and Reward Coverage}
\label{subsec:planner_control}

\noindent\textbf{Training the \emph{Planner}.}
In a matched control that changes only whether the \emph{Planner} is trained, a
trained \emph{Planner} improves both target availability and the downstream
\emph{Editor}. It raises the feasible-source rate, the share of source images
with at least one gate-passing candidate, from $12.5\%$ to $20.3\%$ ($+62.4\%$
relative) and the best \emph{Critic} score by $0.06$, and the downstream
\emph{Editor} gains $+0.10\pm0.02$ SD on ImgEdit over three training seeds. This
shows that proposal quality, and not only target selection, drives
self-distillation. A blinded audit finds correctly grounded proposals for 58 of
64 sources ($\kappa=0.82$; Appendix~\ref{sec:supp_planner_audit}).

\noindent\textbf{Broader reward coverage.}
A four-type verifier-bank model retains the isolation improvement
($+25.8\%$) and reaches a similar overall gain ($+5.3\%$ vs.\ $+5.5\%$
for the primary isolation model; Table~\ref{tab:self_distillation_type_ablation}),
but its composition gain is smaller ($+3.2\%$ vs.\ $+8.4\%$). Covering more
edit types therefore does not compound the benefit of the aligned reward.

\section{Limitations}
\label{subsec:scope}
\method{} amplifies successful behavior that already has nonzero support in the
editor's sampling distribution. Under independent draws with valid-target
probability $p$, the chance of finding a target among $K$ candidates is
$1-(1-p)^K$, so increasing $K$ cannot recover an edit when $p=0$. Gains are also
concentrated where success is internally checkable: on adjustment edits, the
tested internal signals recover at most $30\%$ of the oracle headroom
(Appendix~\ref{sec:supp_verifiability}). The approach requires access to the
editor's multimodal features and latents, and our evidence covers two
multimodal diffusion editors. Evaluation relies on automatic judges, and published
results use their own protocols. Finally, the primary result is a single adaptation pass over one source pool.
Independent human assessment, matched random, scalar, and learned-reward
controls at the primary training budget, and multi-round learning curves would
further strengthen claims about verification quality and sustained
self-improvement.

%% file: tables/transfer_benchmark_results.tex
\begin{table}[!t]
\centering
\caption{\textbf{Transfer of \method{} to GEdit-Bench and Complex-Edit.}
We evaluate the same checkpoint as in Table~\ref{tab:main_results} without any
further adaptation. \method{} improves every reported metric over its
\qwenedit{} base, raising the GEdit-Bench overall score from
$7.39$$\rightarrow8.31$ ($+12.4\%$) and the Complex-Edit overall score from
$8.77\rightarrow8.97$ ($+2.3\%$), with the largest Complex-Edit gains on
perceptual quality ($+4.9\%$) and identity preservation ($+1.8\%$). This
suggests that the verified targets improve editing quality beyond the edit
types emphasized during training. Scores are on the $0$--$10$ scale;
GEdit-Bench reports Semantic Consistency, Perceptual Quality, and Overall
VIEScore~\cite{ku2024viescore}; Complex-Edit reports Instruction Following,
Identity Preservation, Perceptual Quality, and Overall, the mean of the three
metrics. Results of other methods~\cite{liu2025step1xedit,yang2025complexedit,li2025uniworldv2} are reported under their own protocols. $\Delta$ Improvement is the relative improvement over the base.
\extrw{}UniWorld uses an external multimodal-LLM reward model during post-training
and reports $7.54\rightarrow7.76$ ($+2.9\%$) over its own \qwenedit{} base on
GEdit-Bench.}
\label{tab:transfer_benchmark_results}
\small
\setlength{\tabcolsep}{2.6pt}
\begin{tabular}{@{}lccccccc@{}}
\toprule
& \multicolumn{3}{c}{\textbf{GEdit-Bench}} & \multicolumn{4}{c}{\textbf{Complex-Edit}} \\
\cmidrule(lr){2-4}\cmidrule(l){5-8}
\textbf{Model} & \textbf{\shortstack{Semantic\\Consistency}} & \textbf{\shortstack{Perceptual\\Quality}} & \textbf{Overall} & \textbf{\shortstack{Instruction\\Following}} & \textbf{\shortstack{Identity\\Preservation}} & \textbf{\shortstack{Perceptual\\Quality}} & \textbf{Overall} \\
\midrule
\tgroup{8}{Image editing models}
GPT-4o & 7.74 & \textbf{8.13} & 7.49 & 9.29 & 7.51 & \textbf{9.47} & 8.76 \\
Imagen3 & -- & -- & -- & 7.56 & 6.55 & 7.67 & 7.26 \\
SeedEdit & 7.22 & 7.89 & 6.98 & 8.49 & 6.91 & 8.74 & 8.04 \\
Step1X-Edit~\cite{liu2025step1xedit} & 7.13 & 7.00 & 6.44 & -- & -- & -- & -- \\
Step1X-Edit v1.1~\cite{liu2025step1xedit} & 7.66 & 7.35 & 6.97 & -- & -- & -- & -- \\
Gemini 2.0 Flash & 6.87 & 7.44 & 6.51 & -- & -- & -- & -- \\
OmniGen & 5.88 & 5.87 & 5.01 & 6.25 & 6.42 & 7.54 & 6.74 \\
AnyEdit & 3.05 & 5.88 & 2.85 & 1.60 & 8.15 & 7.25 & 5.67 \\
UltraEdit~\cite{zhao2024ultraedit} & -- & -- & -- & 6.56 & 5.93 & 7.29 & 6.59 \\
MagicBrush~\cite{zhang2023magicbrush} & 4.52 & 6.37 & 4.19 & -- & -- & -- & -- \\
InstructPix2Pix~\cite{brooks2023instructpix2pix} & 3.30 & 6.19 & 3.22 & -- & -- & -- & -- \\
\midrule
\tgroup{8}{Reward post-trained methods}
UniWorld-Qwen-Edit\extrw~\cite{li2025uniworldv2} & 8.36 & 7.87 & 7.76 & 9.73 & 9.08 & 7.61 & 8.81 \\
UniWorld-V2\extrw~\cite{li2025uniworldv2} & 8.39 & 8.02 & 7.83 & -- & -- & -- & -- \\
\midrule
Qwen-Image-Edit-2509~\cite{wu2025qwenimage} & 8.11 & 7.10 & 7.39 & 9.69 & 9.02 & 7.60 & 8.77 \\
\oursrow \method{} (ours) & \textbf{9.05} & 7.92 & \textbf{8.31} & \textbf{9.75} & \textbf{9.18} & 7.97 & \textbf{8.97} \\
\oursrow \relrow & \uprel{11.6} & \uprel{11.5} & \uprel{12.4} & \uprel{0.7} & \uprel{1.8} & \uprel{4.9} & \uprel{2.3} \\
\bottomrule
\end{tabular}
\end{table}

%% file: tables/step1x_transfer.tex
\begin{table}[!t]
\centering
\caption{\textbf{\method{} generalizes to a second editor, Step1X-Edit.}
We apply the same procedure to Step1X-Edit using its own VLM and VAE features,
without any \qwenedit{} weights. Self-distillation raises the overall GEdit-Bench
score from $6.69\rightarrow7.24$ ($+8.2\%$) and the overall ImgEdit score from
$3.86\rightarrow4.16$ ($+7.8\%$). This shows that internal verification is not
specific to a single editor. $\Delta$ Improvement is the relative improvement over the base. The overall gains are $+0.55\pm0.05$ SD over three training
seeds for GEdit-Bench and $+0.30\pm0.06$ s.e.\ for ImgEdit.}
\label{tab:step1x_transfer}
\small
\setlength{\tabcolsep}{5pt}
\begin{tabular}{@{}lcccc@{}}
\toprule
& \multicolumn{3}{c}{\textbf{GEdit-Bench}} & \textbf{ImgEdit} \\
\cmidrule(lr){2-4}\cmidrule(l){5-5}
\textbf{Model} & \textbf{\shortstack{Semantic\\Consistency}} & \textbf{\shortstack{Perceptual\\Quality}} & \textbf{Overall} & \textbf{Overall} \\
\midrule
Step1X-Edit~\cite{liu2025step1xedit} & 7.07 & 7.58 & 6.69 & 3.86 \\
\oursrow \method{} (ours) & \textbf{7.84} & \textbf{8.05} & \textbf{7.24} & \textbf{4.16} \\
\oursrow \relrow & \uprel{10.9} & \uprel{6.2} & \uprel{8.2} & \uprel{7.8} \\
\bottomrule
\end{tabular}
\end{table}

%% file: tables/target_selection.tex
\begin{table}[!t]
\centering
\caption{\textbf{The internal reward ranks candidate edits far better than a naive
composite.} We measure how often each signal selects the stronger edit in a
held-out pair ($30$ pairs per family; chance $=0.50$). A naive product of all
reward components collapses on near-ceiling families (e.g., $0.13$ on action
change) because it over-weights preservation, whereas the reward used for target
selection reaches $0.85$ pooled accuracy. This ranking ability allows
self-distillation targets to be mined without an external judge.}
\label{tab:supp_target_selection}
\small
\setlength{\tabcolsep}{6pt}
\begin{tabular}{@{}lcccc@{}}
\toprule
 & \textbf{Naive} & \textbf{VLM} & \textbf{CEPR} & \textbf{Reward} \\
\textbf{Family} & \textbf{composite} & \textbf{success} & \textbf{edit} &
\textbf{(ours)} \\
\midrule
Action change      & $0.13$ & $0.90$ & $0.87$ & $0.87$ \\
Color change       & $0.17$ & $0.80$ & $0.87$ & $0.87$ \\
Compose            & $0.20$ & $0.97$ & $0.96$ & $0.97$ \\
Material change    & $0.27$ & $0.73$ & $0.80$ & $0.80$ \\
Object addition    & $0.33$ & $0.83$ & $0.87$ & $0.87$ \\
Object removal     & $0.60$ & $0.47$ & $0.57$ & $0.60$ \\
Object replacement & $0.87$ & $1.00$ & $0.97$ & $0.97$ \\
\midrule
\oursrow \textbf{Pooled} & $0.37$ & $0.81$ & $0.84$ & $\mathbf{0.85}$ \\
\bottomrule
\end{tabular}
\end{table}

%% file: figures/charts/ablation_summary.tex
% Three aligned native PGFPlots panels sharing one size and style:
% (a) gate validation, (b) best-of-K headroom (Extract family), (c) headroom closed per family (Table 1).
\begingroup\chartmainsetup
\begin{tikzpicture}
% ---- (a) Gate validation ----
\begin{axis}[at={(0.105\linewidth,0pt)},anchor=south west,matched main chart,xmin=-.03,xmax=.85,xtick={0,.4,.8},ymin=-.5,ymax=3.5,y dir=reverse,ytick={0,1,2,3},yticklabels={{Valid\\edit},{No-op},{Corrupt},{Wrong\\edit}},yticklabel style={font=\chartsmall},xlabel={Score / acceptance rate},xmajorgrids=false,ymajorgrids=false,legend columns=2,legend style={at={(.5,1.04)},anchor=south,font=\chartsmall,draw=none,/tikz/every even column/.append style={column sep=4pt}}]
\draw[chartInk!45,densely dotted,line width=.5pt] (axis cs:-.03,.5) -- (axis cs:.85,.5);
\draw[chartInk,line width=.6pt] (axis cs:0,0) -- (axis cs:0.65,0);
\draw[chartInk,line width=.6pt] (axis cs:0,1) -- (axis cs:0.6,1);
\draw[chartInk,line width=.6pt] (axis cs:0,2) -- (axis cs:0.46,2);
\draw[chartInk,line width=.6pt] (axis cs:0,3) -- (axis cs:0.39,3);
\addplot[only marks,mark=o,chartInk,mark size=2.2pt] coordinates {(0.65,0) (0.60,1) (0.46,2) (0.39,3)};
\addlegendentry{Scalar reward}
\addplot[only marks,mark=*,chartGreen,mark size=2.2pt] coordinates {(0.505,0) (0,1) (0,2) (0,3)};
\addlegendentry{Gated accept}
\node[right,font=\chartsmall] at (axis cs:0.65,0) {0.65};
\node[above right,text=chartGreen,font=\chartemph] at (axis cs:0.505,0) {0.51};
\node[right,font=\chartsmall] at (axis cs:0.6,1) {0.60};
\node[above right,text=chartGreen,font=\chartemph] at (axis cs:0,1) {0.00};
\node[right,font=\chartsmall] at (axis cs:0.46,2) {0.46};
\node[above right,text=chartGreen,font=\chartemph] at (axis cs:0,2) {0.00};
\node[right,font=\chartsmall] at (axis cs:0.39,3) {0.39};
\node[above right,text=chartGreen,font=\chartemph] at (axis cs:0,3) {0.00};
\end{axis}
\node[font=\charttitle,align=center,text width=.28\linewidth] at ({0.105\linewidth+.5\chartmainaxiswidth},115pt) {(a) Gate validation};
% ---- (b) Best-of-K headroom ----
\begin{axis}[at={(0.435\linewidth,0pt)},anchor=south west,matched main chart,xmin=-.4,xmax=3.4,ymin=3.3,ymax=4.95,ytick={3.4,3.8,4.2,4.6},ymajorgrids=false,xtick={0,1,2,3},xticklabels={1,2,4,8},xticklabel style={font=\chartsmall},xlabel={Samples $K$ (best-of-$K$)},ylabel={ImgEdit Extract score}]
\draw[chartGreen,dashed,line width=.8pt] (axis cs:-.4,4.26) -- (axis cs:3.4,4.26);
\node[above right,text=chartGreen,font=\chartsmall,align=left,inner sep=1pt] at (axis cs:-.4,4.28) {\shortstack[l]{Distilled, single-shot\\4.26 ($+24.9\%$)}};
\addplot[chartBlue,mark=*,mark size=2.2pt] coordinates {(0,3.41) (1,3.89) (2,4.22) (3,4.45)};
\node[right,xshift=2pt,font=\chartsmall] at (axis cs:0,3.41) {3.41};
\node[below right,font=\chartsmall] at (axis cs:1,3.89) {3.89};
\node[below right,font=\chartsmall] at (axis cs:2,4.22) {4.22};
\node[above,font=\chartsmall,yshift=1pt] at (axis cs:3,4.45) {4.45};
\end{axis}
\node[font=\charttitle,align=center,text width=.28\linewidth] at ({0.435\linewidth+.5\chartmainaxiswidth},115pt) {(b) Best-of-$K$ headroom};
% ---- (c) Headroom closed ----
\begin{axis}[at={(0.765\linewidth,0pt)},anchor=south west,matched main chart,xmin=0,xmax=75,ymin=-.7,ymax=10.5,y dir=reverse,xtick={0,25,50},ytick={0,1,2,3,4,5,6,7,8,9.8},yticklabels={Extract,Replace,Style,Action,Compose,Background,Add,Adjust,Remove,\textbf{Overall}},yticklabel style={font=\chartsmall},xticklabel style={font=\chartsmall},xlabel={Headroom closed (\%)},xmajorgrids=true,ymajorgrids=false]
\addplot[xbar,mark=none,bar width=4pt,fill=chartBlue,draw=chartBlue,forget plot] coordinates {(53.46,0)};
\node[right,font=\chartsmall,inner sep=1.5pt] at (axis cs:53.46,0) {53};
\addplot[xbar,mark=none,bar width=4pt,fill=chartBlue,draw=chartBlue,forget plot] coordinates {(50.0,1)};
\node[right,font=\chartsmall,inner sep=1.5pt] at (axis cs:50.0,1) {50};
\addplot[xbar,mark=none,bar width=4pt,fill=chartBlue,draw=chartBlue,forget plot] coordinates {(43.3,2)};
\node[right,font=\chartsmall,inner sep=1.5pt] at (axis cs:43.3,2) {43};
\addplot[xbar,mark=none,bar width=4pt,fill=chartBlue,draw=chartBlue,forget plot] coordinates {(41.9,3)};
\node[right,font=\chartsmall,inner sep=1.5pt] at (axis cs:41.9,3) {42};
\addplot[xbar,mark=none,bar width=4pt,fill=chartBlue,draw=chartBlue,forget plot] coordinates {(35.8,4)};
\node[right,font=\chartsmall,inner sep=1.5pt] at (axis cs:35.8,4) {36};
\addplot[xbar,mark=none,bar width=4pt,fill=chartBlue,draw=chartBlue,forget plot] coordinates {(31.7,5)};
\node[right,font=\chartsmall,inner sep=1.5pt] at (axis cs:31.7,5) {32};
\addplot[xbar,mark=none,bar width=4pt,fill=chartBlue,draw=chartBlue,forget plot] coordinates {(28.6,6)};
\node[right,font=\chartsmall,inner sep=1.5pt] at (axis cs:28.6,6) {29};
\addplot[xbar,mark=none,bar width=4pt,fill=chartBlue,draw=chartBlue,forget plot] coordinates {(23.3,7)};
\node[right,font=\chartsmall,inner sep=1.5pt] at (axis cs:23.3,7) {23};
\addplot[xbar,mark=none,bar width=4pt,fill=chartBlue,draw=chartBlue,forget plot] coordinates {(21.9,8)};
\node[right,font=\chartsmall,inner sep=1.5pt] at (axis cs:21.9,8) {22};
\addplot[xbar,mark=none,bar width=4pt,fill=chartInk,draw=chartInk,forget plot] coordinates {(37.5,9.8)};
\node[right,font=\chartsmall\bfseries,inner sep=1.5pt] at (axis cs:37.5,9.8) {38};
\draw[chartInk!50,densely dotted] (axis cs:0,8.9) -- (axis cs:75,8.9);
\end{axis}
\node[font=\charttitle,align=center,text width=.28\linewidth] at ({0.765\linewidth+.5\chartmainaxiswidth},115pt) {(c) Headroom closed};
\end{tikzpicture}

\endgroup

%% file: sections/6_conclusion.tex
\section{Conclusion}
\label{sec:conclusion}
We introduced \method{}, a self-evolving framework that improves a pretrained
image editor without human-edited targets or an external training-time reward
model. By coupling a \emph{Planner}, an \emph{Editor}, and a fixed internal \emph{Critic} in a
propose--edit--verify--distill loop, and verifying each candidate with
decomposed rubric checks and non-compensatory gates, the editor learns only from
its own verified successes. Our experiments show consistent gains on ImgEdit,
with the largest improvement on object isolation, transfer to GEdit-Bench and
Complex-Edit, and a second successful instantiation on Step1X-Edit. Our analyses
show that verified targets give larger gains than naive self-training, and that
training the \emph{Planner} further improves the \emph{Editor}. This work takes a step toward image editors that improve
themselves from their own verified samples, and suggests future directions in
richer internal verifiers, multi-round self-evolution, and broader edit types.

%% file: appendix/supplementary.tex
% Shared supplementary body; the document root supplies the format and numbering.
% Captions outside floats link to their caption rather than a float anchor.
\captionsetup{hypcap=false}
\setlist[itemize]{leftmargin=*,itemsep=1pt,topsep=2pt,parsep=0pt,partopsep=0pt}
\setlist[enumerate]{leftmargin=*,itemsep=1pt,topsep=2pt,parsep=0pt,partopsep=0pt}

\section{Supplementary Experiments and Details}
\label{sec:supp_overview}

This appendix provides additional details and analyses supporting the main
paper. We present (1) sampling and target-selection diagnostics, (2) extended
reward-signal and headroom studies, (3) the complete training algorithms,
(4) evaluation protocols and self-training comparisons, (5) implementation
details, (6) additional ablations, and (7) additional qualitative results. The primary internal-only model and broader
verifier-bank analyses are distinguished throughout.

\section{Sampling and Target Construction}
\label{sec:supp_self_evolution_analysis}

The method relies on the editor's own stochastic successes being selected by an
aligned internal reward and distilled into a single-shot checkpoint. We analyze
four properties of this pipeline: the base
distribution contains recoverable successful edits; the final model
internalizes those edits rather than relying on test-time selection; and the reward
gate rejects invalid pseudo-labels.
Fig.~\ref{fig:ablation_summary}b visualizes the recoverable headroom
diagnostics, and the following sections report the supporting studies in detail.

\subsection{Naive-Loop Target Distribution}
\begin{figure}[!htbp]
\centering
\input{figures/charts/diagnostic_evidence}
\caption{\textbf{Without verification, naive self-training collects more failures than
usable targets.} We analyze the training manifest of a four-round
self-training loop without feasibility gates. (a) Cumulative accepted,
rejected, and replay rows per round: by the final round, rejected candidates
outnumber accepted ones ($546$ vs.\ $302$). (b) Accepted clean targets per edit
type: color edits dominate the pool, while preservation-sensitive removal and
replacement receive only $3$ and $2$ clean targets ($92\%$ and $96\%$ rejected).
This imbalance explains why the naive loop improves the base editor less than
verified targets do, and motivates verifying each target before distillation.}
\label{fig:diagnostic_evidence}
\end{figure}
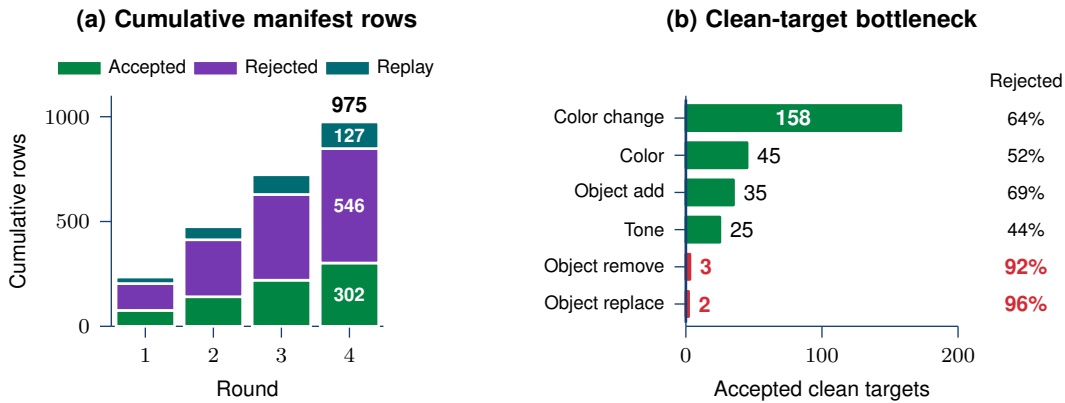

The four-round diagnostic retains 302 accepted, 546 rejected, and 127 replay
records by its final round. Its accepted targets are concentrated in color
edits, with few clean removal and replacement targets. These counts describe
target construction in the naive-loop configuration; they are not a
held-out quality trajectory. Figure~\ref{fig:diagnostic_evidence} visualizes the manifest.

\subsection{Best-of-\texorpdfstring{$K$}{K} Sampling Diagnostics}
Figure~\ref{fig:ablation_summary}b reports the best-of-$K$ headroom on the full ImgEdit extraction family: reward-best
selection raises the score from $3.41$ at $K{=}1$ to $3.89$, $4.22$, and $4.45$ at $K{=}2,4,8$,
showing that the base distribution already contains many successful isolations,
while the self-distilled checkpoint reaches $4.26$ with a single generation.

This analysis motivates reward-verified distillation rather than a best-of-$K$
multi-sample inference procedure. Best-of-$K$ confirms that
the base distribution contains successful hard edits, but it requires multiple
samples and a selector at inference. The reported result instead evaluates the
trained checkpoint in the same single-sample setting as the base, so the
headroom must be internalized into the editor weights.
The analysis also separates sample availability from sample validity: increasing
$K$ makes more successful edits visible, but it also exposes more invalid
candidates. Consequently, the remaining ablations focus on the gate that
decides which candidates are suitable for denoising targets.

\section{Extended Reward-Signal and Headroom Studies}
\label{sec:supp_extended}

This section reports four studies that extend the main paper's evidence.
First, a six-family stochastic probe shows that recoverable best-of-$K$ headroom
is largest for object isolation, the weakest self-verifiable base family
(Sec.~\ref{sec:supp_perfamily_headroom}). Second, on the same
$210$-pair probe used for the reward-gate ablation, the internal reward selects
the better of two candidate edits far more reliably than a naive composite of
its components (Sec.~\ref{sec:supp_target_selection}). Third, an oracle-capture
analysis on the adjust family illustrates a \emph{self-verifiability limitation}:
the method recovers headroom only when a family's success criterion is
internally checkable, which motivates examining verifier quality before training (Sec.~\ref{sec:supp_verifiability}). Fourth, a Step1X-Edit diagnostic
tests whether the same headroom pattern appears on a different editor backbone
(Sec.~\ref{sec:supp_cross_backbone}).

\subsection{Per-Family Recoverable Headroom}
\label{sec:supp_perfamily_headroom}

The largest primary gain occurs in the weakest base family. Table~\ref{tab:supp_perfamily_headroom} examines sampling headroom across six families.
On a $75$-example, five-seed stochastic probe (a separate sample, so its base
scores differ from Table~\ref{tab:main_results}) we draw five independent samples
per instruction and record the reward-best score at $K\in\{1,2,4,5\}$. Object
isolation, the weakest base family ($2.97$), exposes a large $+1.86$
best-of-5 gain. The three near-ceiling families (base $4.60$--$4.62$) gain only $+0.25$
to $+0.28$: action and replacement approach the grader's $5.0$ ceiling
($4.87$--$4.88$) at $K{=}4$, as does style/tone. These families leave less recoverable
headroom than object isolation. Adjust and removal sit in between,
at $+0.59$ and $+0.49$. Weighting the per-family deltas by their true ImgEdit
type counts (covering $525/737$ examples) yields a full-benchmark selection projection of $+0.51$ at best-of-5,
assuming the uncovered families remain unchanged. Whether this selected headroom is
\emph{realizable} by single-shot self-distillation depends on internal
verifiability (Sec.~\ref{sec:supp_verifiability}), so this probe should be read
as a target-selection diagnostic across edit families.

\begin{table}[!tb]
\centering
\caption{\textbf{Object isolation offers the largest recoverable headroom.}
We report the reward-best ImgEdit-style score at each $K$ on a six-family
probe, where $\Delta_{5}$ is the best-of-5 gain over the base ($K{=}1$) mean.
Object isolation gains $+1.86$, $6$--$7\times$ more than the near-ceiling
families, which approach the grader's $5.0$ ceiling. This makes object
isolation the most promising target for self-distillation, which we adopt as
the primary instantiation. Deltas are computed before rounding the means.}
\label{tab:supp_perfamily_headroom}

\small
\setlength{\tabcolsep}{6pt}
\begin{tabular}{@{}lccccc@{}}
\toprule
\textbf{Edit family} & \textbf{Base} & \textbf{$K{=}2$} & \textbf{$K{=}4$} &
\textbf{$K{=}5$} & \textbf{$\Delta_{5}$} \\
\midrule
Object isolation   & $2.97$ & $4.15$ & $4.68$ & $4.82$ & $\mathbf{+1.86}$ \\
Adjust (attribute) & $4.04$ & $4.41$ & $4.53$ & $4.62$ & $+0.59$ \\
Object removal     & $3.92$ & $4.23$ & $4.39$ & $4.42$ & $+0.49$ \\
Action change      & $4.60$ & $4.84$ & $4.88$ & $4.89$ & $+0.28$ \\
Object replacement & $4.62$ & $4.86$ & $4.88$ & $4.89$ & $+0.28$ \\
Style / tone       & $4.61$ & $4.83$ & $4.87$ & $4.87$ & $+0.25$ \\
\bottomrule
\end{tabular}

\end{table}

\subsection{Reward-Verified Target Selection}
\label{sec:supp_target_selection}

Self-distillation must not only accept a valid edit (Sec.~\ref{sec:supp_gate})
but also pick the \emph{better} of two valid candidates as the denoising target.
Table~\ref{tab:supp_target_selection} (main paper) evaluates this on the same $210$-pair
probe: each pair contrasts a stronger and a weaker edit for one instruction, and
the reward must select the stronger. A naive product of all reward components
picks correctly only $36.7\%$ of the time --- below chance on the near-ceiling
families (action $13\%$, color $17\%$, compose $20\%$) because it over-weights
the preservation term and penalizes large but legitimate edits. The verifier-bank reward balances edit and preservation terms after
applicable feasibility checks, raising pooled selection accuracy to $84.8\%$ (chance $=50\%$), with $80$--$97\%$ on six families and $60\%$ on removal. This is the ranking behavior required to mine
self-distillation targets without an external judge.

\subsection{The Self-Verifiability Boundary}
\label{sec:supp_verifiability}

The per-family probe shows the adjust family carries a real $+0.52$ oracle
headroom on a controlled $12$-instance subset, yet the distilled model improves
it by only $+0.14$ ($+3.2\%$), consistent with the per-type discussion in
Sec.~\ref{sec:supp_type_ablation}.
Table~\ref{tab:supp_capture} explains the gap. We ask whether \emph{any} internal signal can select the
oracle-best adjust sample from six candidates. Best-of-6 oracle selection
recovers the full $+0.52$; but a VLM edit-success judge recovers only $+0.16$
($30\%$ of the gap), a diff-region crop of the same judge matches it, a pairwise
VLM comparison recovers $+0.05$, and a deterministic computer-vision
color-matching tool recovers $+0.02$. CEPR-based semantic signals are actively
\emph{negative} ($-0.18$, $-0.26$): they rank perceptually worse samples higher.
The pattern suggests a limitation of the tested internal signals on this
probe. Multiple plausible adjust outputs satisfy the instruction, while the
tested checks recover little of the oracle's perceptual-quality advantage.
Extraction, removal, and addition offer more explicit after-state criteria,
such as a clean background, object absence, or object presence.
Table~\ref{tab:supp_capture} motivates measuring selector quality before
training; it does not establish that a family is universally unverifiable.

\begin{table}[!tb]
\centering
\caption{\textbf{Internal signals recover only part of the oracle headroom on
adjustment edits.} We test whether any internal signal can pick the oracle-best
of six adjustment samples. Capture fraction is
$(\text{signal-best}-\text{mean})/(\text{oracle}-\text{mean})$ over $12$
instances $\times$ six seeds, and recovered score expresses the same quantity in
ImgEdit-style points. The best internal signal recovers only $30\%$ of the
oracle headroom, and semantic-embedding signals are anti-correlated with
perceptual quality. Self-distillation is therefore most effective on edit types
whose success criterion can be checked internally.}
\label{tab:supp_capture}

\small
\setlength{\tabcolsep}{7pt}
\begin{tabular}{@{}lcc@{}}
\toprule
\textbf{Selection signal} & \textbf{Capture} & \textbf{Recovered} \\
 & \textbf{(fraction)} & \textbf{(score pts)} \\
\midrule
Oracle (best-of-6)        & $1.00$  & $+0.52$ \\
VLM edit-success judge      & $0.30$  & $+0.16$ \\
VLM judge, diff-region crop & $0.30$  & $+0.16$ \\
Pairwise VLM comparison     & $0.09$  & $+0.05$ \\
CV color-match tool        & $0.04$  & $+0.02$ \\
CEPR semantic margin        & $-0.34$ & $-0.18$ \\
CEPR margin, diff-region crop & $-0.50$ & $-0.26$ \\
\bottomrule
\end{tabular}

\end{table}

\subsection{Cross-Backbone Diagnostic: Step1X-Edit}
\label{sec:supp_cross_backbone}

The observed headroom pattern (Sec.~\ref{sec:supp_perfamily_headroom}) and the
self-verifiability boundary (Sec.~\ref{sec:supp_verifiability}) should not be
specific to one editor. We therefore repeat the diagnostic on
Step1X-Edit~\cite{liu2025step1xedit}, using its own Qwen2.5-VL feature
interface and DiT decoder. This diagnostic does not import
\qwenedit{} weights~\cite{wu2025qwenimage}.
We compare both editors' base scores per family on the same probe under the
identical GPT-4o protocol.

The results show that object isolation is the single weakest family on
\emph{both} backbones, and it is markedly weaker on Step1X-Edit ($1.64$ vs.\
$2.97$). The ordering of the remaining families is also largely preserved ---
removal is the second-weakest on both ($3.29$ and $3.92$), while action,
replacement, and style sit near the grader ceiling. Object isolation has a
well-defined, internally checkable success criterion (the requested object
isolated on a clean background), placing it squarely inside the self-verifiable
region of Sec.~\ref{sec:supp_verifiability}. We next measure whether these weak
families also contain recoverable best-of-$K$ headroom.

Table~\ref{tab:supp_step1x_diagnostic} summarizes the full Step1X diagnostic in
one view. This five-seed GPT-4o probe is separate from the three-training-seed
GEdit-Bench and ImgEdit experiments reported in
the main paper; their scores should not be compared across
protocols. Both weak families contain best-of-5 oracle headroom, but recovery
tracks reward quality: removal has a stronger reward--quality correlation
($0.64$), so the reward pick captures $61\%$ of oracle headroom and the
self-distilled LoRA recovers $57\%$ into a single pass. Object isolation has weaker
reward correlation ($0.16$), and the distilled gain is correspondingly smaller.

\begin{table}[!tb]
\centering
\caption{\textbf{Headroom recovery on Step1X-Edit follows reward quality.}
We repeat the headroom diagnostic on the two weakest Step1X-Edit families using
five-seed GPT-4o means. Oracle is the grader-best of five samples, Reward pick is
the sample chosen by the fixed internal \emph{Critic} before training, and Self-dist.\
is the single-pass rank-16 LoRA trained on reward-selected targets. For removal,
where the reward correlates well with quality ($0.64$), the reward pick captures
$61\%$ of the oracle headroom and self-distillation retains $57\%$; for object
isolation (correlation $0.16$) recovery is smaller, suggesting that better
internal rewards translate into larger distilled gains.}
\label{tab:supp_step1x_diagnostic}

\small
\setlength{\tabcolsep}{3.2pt}
\begin{tabular}{@{}lcccccc@{}}
\toprule
\textbf{Family} & \textbf{Base} & \textbf{Oracle} &
\textbf{Reward} & \textbf{Reward} &
\textbf{Self} & \textbf{Self} \\
 &  & \textbf{best-of-5} & \textbf{pick} & \textbf{capt.} &
\textbf{dist.} & \textbf{capt.} \\
\midrule
Object removal    & $3.29$ & $4.27$ & $\mathbf{3.89}$ & $61\%$ & $3.85$ & $57\%$ \\
Object isolation  & $1.64$ & $2.39$ & $1.86$          & $30\%$ & $1.80$ & $21\%$ \\
\bottomrule
\end{tabular}

\end{table}

\section{Algorithms}
\label{sec:supp_algorithms}

Algorithm~\ref{alg:supp_loop} formalizes the general round-based distillation loop,
and Algorithm~\ref{alg:supp_reward} specifies the gated internal reward used to
select self-distillation targets.

\begin{suppalgorithm}{Reward-verified self-distillation loop.}
\label{alg:supp_loop}
\textbf{Input:} unlabeled image pool $\mathcal{X}$, base editor $E_0$, \emph{Planner}
$P_0$, fixed internal \emph{Critic} $C$, rounds $T$, candidates per proposal $K$.

\textbf{For each round $t=0,\ldots,T-1$:}
\begin{enumerate}
\item Sample a source shard $\mathcal{X}_t\subset\mathcal{X}$, optionally
stratified toward edit families with low base scores.
\item For every source image $x\in\mathcal{X}_t$, let the \emph{Planner} emit a natural
language instruction $c$ and structured specification $s$ containing edit type,
source/target entities, required after-state, forbidden after-state, target
region, and preservation constraints.
\item Generate $K$ \emph{Editor} samples $\{y_i\}_{i=1}^{K}$ with the current \emph{Editor}
$E_t(x,c)$ under matched generation settings.
\item Score each candidate with the internal \emph{Critic}:
$R_i=C(x,c,s,y_i)$, retaining all component scores, gates, and ranks.
\item If at least one candidate is feasible, select
$y^\star=\arg\max_i R_i$ over gate-passing candidates and write
$(x,c,s,y^\star,R^\star)$ to the self-distillation manifest.
\item Train an \emph{Editor} adapter on accepted records plus replay/identity records,
using reward-weighted denoising SFT. Train the \emph{Planner} only from band-pass
proposal traces, where neither all candidates fail nor all candidates pass.
\item Promote the \emph{Editor} and \emph{Planner} checkpoints for the next round. Rejected
candidates are retained for analysis and excluded from SFT targets.
\end{enumerate}
\textbf{Output:} the final editor $E_T$ and a manifest containing
source images, instructions, candidates, rewards, gates, and selected targets.
\end{suppalgorithm}

\begin{suppalgorithm}{Gated internal reward.}
\label{alg:supp_reward}
\textbf{Input:} source image $x$, instruction $c$, structured edit
specification $s$, candidate edit $y$, exposed Qwen image/text features
$\phi_Q(\cdot)$ and $\tau_Q(\cdot)$, VAE latents $z_x,z_y$.

\begin{enumerate}
\item Compute prompt support
$a(I,p)=\frac{1}{2}\left(1+\cos(\phi_Q(I,p),\tau_Q(p))\right)$ and
source-to-edit gains $d_p(y,x)=a(y,p)-a(x,p)$ for required, forbidden,
counterfactual, and preservation prompts.
\item Evaluate source grounding $S$, required after-state $A$, forbidden
after-state $F$, explicit preservation $P_{\mathrm{rub}}$, contrastive edit
$E_{\mathrm{ctr}}$, taxonomy edit $E_{\mathrm{tax}}$, scalar
$P_{\cepr}$ preservation, and latent locality/validity $V$.
\item When an edit type has an editor-side verifier-bank check, evaluate the
yes/no VLM gate $J$. Unsupported checks are omitted rather than filled with a
neutral hallucinated score.
\item Form edit and preservation aggregates
\begin{equation}
  E=\gem(A,F,E_{\mathrm{ctr}},E_{\mathrm{tax}}),\qquad
  P=\gem(P_{\cepr},P_{\mathrm{rub}}).
\end{equation}
\item Apply the non-compensatory gate
\begin{equation}
  G=G_{\mathrm{checks}}\,\indic[\sqrt{EP}\geq\tau_R],\qquad
  G_{\mathrm{checks}}=\textstyle\prod_{u\in\mathcal{U}}\indic[u\geq\tau_u].
\end{equation}
where $\mathcal{U}$ contains the supported checks from
$\{S,A,F,P_{\mathrm{rub}},E_{\mathrm{ctr}},E_{\mathrm{tax}},
P_{\cepr},V,J\}$ after omitting unsupported terms.
\item Return $R=G\sqrt{EP}$. If any required check fails, the candidate is
infeasible and cannot be selected, even if another score is high.
\end{enumerate}
\end{suppalgorithm}

\section{Evaluation Protocols}
\label{sec:supp_protocols}

All deltas for our adapted editors are matched against the same base within
the same protocol. Literature values retain their original source protocols. We use ImgEdit~\cite{ye2025imgedit}, Complex-Edit~\cite{yang2025complexedit}, and
GEdit-Bench~\cite{liu2025step1xedit} scored with VIEScore~\cite{ku2024viescore};
absolute scores are interpreted only within their corresponding scorer and
benchmark protocol.

The separation is important for interpreting the results. ImgEdit is the typed
edit benchmark used to locate the weakest family and report the main in-domain
gain. The 11-task GEdit-Bench study uses a different judge from ImgEdit.
Complex-Edit is a broader compound-edit evaluation
that measures whether the checkpoint preserves instruction following, identity,
and perceptual quality. The reward-gate and per-type analyses
characterize pseudo-label quality and reward coverage.

\subsection{Self-Training Procedures and Signal Sources}
\label{sec:supp_score_accounting}

Table~\ref{tab:supp_score_accounting} compares the reward-verified procedure
with an ungated self-training loop and a pairwise DPO diagnostic. Each row
compares a checkpoint with its own base under one benchmark protocol; absolute
scores are not compared across judges.

\begin{table}[!tb]
\centering
\caption{\textbf{Verified targets give the largest gains over naive self-training and
DPO.} We compare an ungated self-training loop, pairwise DPO on generated
targets, and our reward-verified self-distillation, each against the
\qwenedit{} base under the same protocol. \method{} yields the largest gains
($+5.5\%$ on ImgEdit and $+12.4\%$ on GEdit-Bench); the naive loop gains
$+2.0\%$ on ImgEdit and $+10.5\%$ on GEdit-Bench, and DPO is roughly neutral
($+0.7\%$). $\Delta$ Improvement is the relative change over the
base. The naive-loop GEdit-Bench evaluation uses a different scorer (base $7.33$
vs.\ $7.39$), and the DPO diagnostic uses a smaller diagnostic split of ImgEdit.}
\label{tab:supp_score_accounting}
\small
\setlength{\tabcolsep}{5pt}
\begin{tabular}{@{}lcccccc@{}}
\toprule
& \multicolumn{3}{c}{\textbf{ImgEdit}} & \multicolumn{3}{c}{\textbf{GEdit-Bench}} \\
\cmidrule(lr){2-4}\cmidrule(l){5-7}
\textbf{Procedure} & \textbf{Base} & \textbf{Adapted} & \textbf{$\Delta$ Impr.} &
\textbf{Base} & \textbf{Adapted} & \textbf{$\Delta$ Impr.} \\
\midrule
Naive self-training loop (ungated) & 4.36 & 4.45 & \uprel{2.0} & 7.33 & 8.10 & \uprel{10.5} \\
Pairwise DPO~\cite{rafailov2023dpo,wallace2024diffusiondpo} & 4.56 & 4.59 & \uprel{0.7} & -- & -- & -- \\
\oursrow \method{} (reward-verified) & 4.36 & \textbf{4.60} & \textbf{\uprel{5.5}} & 7.39 & \textbf{8.31} & \textbf{\uprel{12.4}} \\
\bottomrule
\end{tabular}
\end{table}

Table~\ref{tab:supp_signal_sources} separates training-time reward signals, analysis signals, and final evaluation judges. This separation is important because the primary method is trained without querying an external
reward model, while external judges are still needed for benchmark measurement.

\begin{table}[!tb]
\centering
\caption{\textbf{\method{} trains only on signals available inside the editor.}
We list the signal used at each stage and whether it involves an external
model. Target selection for the primary model uses editor-exposed features, VAE
latents, and image statistics only. External judges are used solely for
benchmark evaluation.}
\label{tab:supp_signal_sources}

\small
\setlength{\tabcolsep}{3.5pt}
\setlength{\tabcolsep}{4pt}
\begin{tabular}{@{}L{0.21\linewidth}L{0.43\linewidth}cL{0.20\linewidth}@{}}
\toprule
\textbf{Stage} & \textbf{Signal} & \textbf{\shortstack{External\\model}} & \textbf{Role} \\
\midrule
Source pool & Unlabeled off-benchmark images; \emph{Planner}-generated instructions and specifications & \xmark & Training inputs \\
\oursrow Primary target selection & Editor-exposed Qwen image/text features, VAE latents, image statistics & \xmark & Training reward \\
Verifier-bank analysis & Editor-side VLM yes/no log-probabilities; detector for the addition variant only & detector only & Analysis \\
Benchmark evaluation & ImgEdit GPT scoring, GPT-4.1 VIEScore, Complex-Edit metrics & judge & Evaluation only \\
\bottomrule
\end{tabular}

\end{table}

\section{Implementation Details}
\label{sec:supp_implementation}

The base editor is \texttt{Qwen-Image-Edit-2509}. The round-level analysis uses
four rounds, 128 records per round, two candidates per proposal, and a rank-16
editor LoRA~\cite{hu2022lora} trained for 128 steps per round. The
reward-verified run removes rejected candidates from SFT, uses four candidates
per proposal, seeds sources for underrepresented edit types, and trains an
attention-only rank-16, alpha-16 LoRA adapter anchored to the warm-start
checkpoint. Table~\ref{tab:supp_impl_settings} summarizes the implementation
settings used for the reported experiments.

\begin{table}[!tb]
\centering
\caption{\textbf{Implementation details of \method{}.}
We list the generation, adapter, round-level, \emph{Planner}, and verifier settings
used for the reported experiments. Every reported gain compares the adapted
model with the same base, benchmark, and scorer protocol. Rejected candidates
are kept for analysis but never used as denoising-SFT targets.}
\label{tab:supp_impl_settings}

\small
\setlength{\tabcolsep}{6pt}
\begin{tabular}{@{}L{0.36\linewidth}L{0.56\linewidth}@{}}
\toprule
\textbf{Setting} & \textbf{Value} \\
\midrule
\tgroup{2}{Generation}
Target / conditioning resolution & $1024$ px / $384$ px, aspect ratio preserved \\
Max.\ sequence length & $512$ \\
Scheduler shift / mode scale & $3.0$ / $1.29$ \\
Candidates per proposal & $4$ (round-level analysis: $2$) \\
\midrule
\tgroup{2}{Editor adapter (self-distillation run)}
LoRA placement & Attention QKV and output projections \\
Rank / alpha / dropout & $16$ / $16$ / $0$ \\
Optimizer, precision & AdamW, bf16 \\
Learning rate & $10^{-4}$ \\
SFT steps & $400$ \\
Batch size / grad.\ accumulation & $1$ / $1$ \\
Seed & $123$ \\
Initialization & Warm-start checkpoint (anchored) \\
\midrule
\tgroup{2}{Round-level analysis}
Rounds / records per round & $4$ / $128$ \\
\emph{Editor} LoRA rank / steps per round & $16$ / $128$ \\
Learning rate & $5{\times}10^{-5}$ \\
\midrule
\tgroup{2}{Planner adapter}
Rank / alpha & $16$ / $32$ \\
Learning rate / steps & $10^{-5}$ / $16$ \\
Targets & Reward-weighted normalized JSON \\
\midrule
\tgroup{2}{Verifier thresholds}
Edit / preservation / validity / reward & $0.40$ / $0.20$ / $0.50$ / $0.30$ \\
Taxonomy / source grounding & $0.35$ / $0.42$ \\
Required / forbidden after-state & $0.45$ / $0.42$ \\
Rubric preservation / reward & $0.30$ / $0.28$ \\
Conservative-region difference & $18$ \\
Detector box / text / score$^{*}$ & $0.22$ / $0.18$ / $0.45$ \\
\bottomrule
\multicolumn{2}{@{}l}{\footnotesize $^{*}$Addition verifier-bank variant only.}
\end{tabular}

\end{table}

\paragraph{Self-distillation manifest.}
Each accepted record stores the source-image identifier, instruction, structured
edit specification, candidate identifiers, reward components, gate decisions,
selected target, and replay flag. The manifest links every accepted denoising
target to its source instruction and keeps rejected candidates separate from
SFT.

\section{Additional Ablation Studies}
\label{sec:supp_ablations}

\subsection{Reward Gate Components}
\label{sec:supp_gate}

The reward is designed as a selector, not merely as a ranking score. The
required behavior is therefore two-dimensional: a useful reward must accept
valid edits while rejecting no-op, corrupt, and wrong-edit candidates. The
component ablation reports the verifier-bank operating point: the full gate accepts $50.5\%$ of valid edits while keeping all
three negative-control acceptances at $0.0\%$. The only ablated variant that
admits corrupted edits removes the editor-side VLM check, so that check remains
part of the verifier-bank setting. We therefore prioritize zero negative-control
acceptance over looser valid-edit recall, because accepted candidates become
denoising targets in subsequent self-distillation.

\input{tables/reward_gate_ablation}

\subsection{Per-Type Self-Distillation}
\label{sec:supp_type_ablation}

The per-type analysis shows that self-distillation follows base-model headroom:
the aligned object-isolation checkpoint gives the best overall ImgEdit gain
because it directly targets the weakest base family. The broader verifier-bank
model keeps the object-isolation improvement but redistributes smaller gains
across secondary families, yielding a slightly lower aggregate gain. The
per-task GEdit-Bench analysis below reports behavior under a different
evaluation scorer.

\paragraph{Verifier-bank confidence signal.}
Verbalized confidence takes only two realized values and ties 93\% of candidate
pairs in the diagnostic. Yes/no token log-probabilities reduce ties to 7\%;
their correlation with the evaluation judge is 0.32, versus 0.06 for verbalized
confidence. These are descriptive measurements of the broader verifier-bank
signal, not an independent human-alignment benchmark.

\paragraph{Secondary-family effects.}
Outside object isolation, the primary checkpoint gains between $+0.13$ and
$+0.34$ in each of the other eight families (unweighted mean $+0.17$).
A 42-item off-target probe has mean change 0.11.
Across the nine families, the descriptive correlation between base score and
gain is $r=-0.94$. The small number of families and unequal headroom prevent
interpreting this correlation as a universal prediction rule.

\input{tables/self_distillation_type_ablation}

\subsection{\texorpdfstring{\emph{Planner}}{Planner} Proposal Audit}
\label{sec:supp_planner_audit}
A blinded 64-source audit records 63 valid emissions, 61 complete
specifications, and 60 unique proposals. Grounding is correct for 58/64
sources, with inter-rater agreement $\kappa=0.82$. All 13 scheduled proposal
types occur; this schedule differs from the nine ImgEdit benchmark families.
These measurements assess proposal grounding and coverage, rather than human
preferences over the final edited images. The matched \emph{Planner} training
comparison is reported in Section~\ref{subsec:planner_control}.

\subsection{Per-Task GEdit Transfer}
\label{sec:supp_gedit}

The aggregate GEdit improvement averages over 11 tasks, and Table~\ref{tab:supp_gedit_tasks}
shows the task-level scores. Gains are strongest for style change, human
photo editing, and tone transfer, among the tasks with the lowest base scores, and weakest
for color alteration; no task regresses. This is consistent with the main-paper
result: the checkpoint improves most where the base editor is weakest.

\begin{table}[!tb]
\centering
\caption{\textbf{\method{} improves every GEdit-Bench task.}
We report the Semantic Consistency, Perceptual Quality, and Overall VIEScore of
\qwenedit{} (base) and \method{} for each of the 11 tasks, and the Mean row
matches Table~\ref{tab:transfer_benchmark_results}. Gains are largest on style
change, human photo editing, and tone transfer, which are among the tasks with the lowest base scores,
and smallest on color alteration; no task regresses.}
\label{tab:supp_gedit_tasks}

\small
\setlength{\tabcolsep}{6pt}
\begin{tabular}{@{}lcccccc@{}}
\toprule
& \multicolumn{2}{c}{\textbf{Semantic Consistency}} & \multicolumn{2}{c}{\textbf{Perceptual Quality}} & \multicolumn{2}{c}{\textbf{Overall}} \\
\cmidrule(lr){2-3}\cmidrule(lr){4-5}\cmidrule(l){6-7}
\textbf{GEdit-Bench task} & \textbf{Base} & \textbf{Ours} & \textbf{Base} & \textbf{Ours} & \textbf{Base} & \textbf{Ours} \\
\midrule
Background change & 8.51 & \textbf{9.33} & 7.13 & \textbf{7.83} & 7.61 & \textbf{8.41} \\
Color alteration & 8.36 & \textbf{8.96} & 7.21 & \textbf{7.62} & 7.58 & \textbf{8.13} \\
Material alteration & 8.25 & \textbf{9.09} & 5.46 & \textbf{6.29} & 6.57 & \textbf{7.44} \\
Motion change & 8.13 & \textbf{8.81} & 8.42 & \textbf{9.04} & 8.10 & \textbf{8.78} \\
Human photo editing & 7.19 & \textbf{8.27} & 6.82 & \textbf{8.04} & 6.86 & \textbf{8.02} \\
Style change & 7.45 & \textbf{8.79} & 4.97 & \textbf{6.28} & 5.94 & \textbf{7.31} \\
Subject addition & 8.74 & \textbf{9.75} & 7.48 & \textbf{8.26} & 7.91 & \textbf{8.83} \\
Subject removal & 8.70 & \textbf{9.97} & 7.28 & \textbf{7.97} & 7.77 & \textbf{8.77} \\
Subject replacement & 8.50 & \textbf{9.31} & 7.61 & \textbf{8.11} & 7.87 & \textbf{8.55} \\
Text change & 8.17 & \textbf{8.89} & 8.09 & \textbf{9.06} & 7.94 & \textbf{8.83} \\
Tone transfer & 7.22 & \textbf{8.40} & 7.62 & \textbf{8.62} & 7.23 & \textbf{8.37} \\
\midrule
\textbf{Mean} & 8.11 & \textbf{9.05} & 7.10 & \textbf{7.92} & 7.39 & \textbf{8.31} \\
\bottomrule
\end{tabular}

\end{table}

\section{Additional Qualitative Samples}
\label{sec:supp_qualitative}

\begin{figure}[!htbp]
\centering
\includegraphics[width=\linewidth,height=0.60\textheight,keepaspectratio]{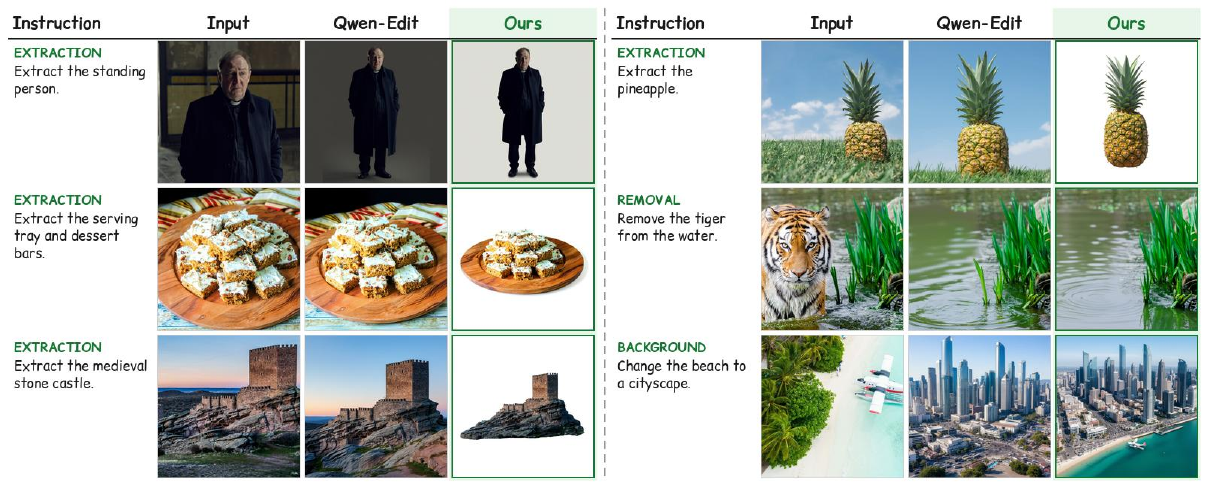}
\caption{\textbf{Additional qualitative results on high-headroom edits.}
Each row shows the instruction, the source image, the \qwenedit{} output, and
our \method{} output on unused ImgEdit examples. The extraction rows use the
object-isolation checkpoint, and the removal and background rows use the final
\method{} checkpoint. Where the baseline fails to realize the requested
operation, \method{} cleanly separates the requested object from its context and
localizes spatially specific removal and background edits.}
\label{fig:supp_qual_high_gain}
\end{figure}

\begin{figure}[!htbp]
\centering
\includegraphics[width=\linewidth,height=0.60\textheight,keepaspectratio]{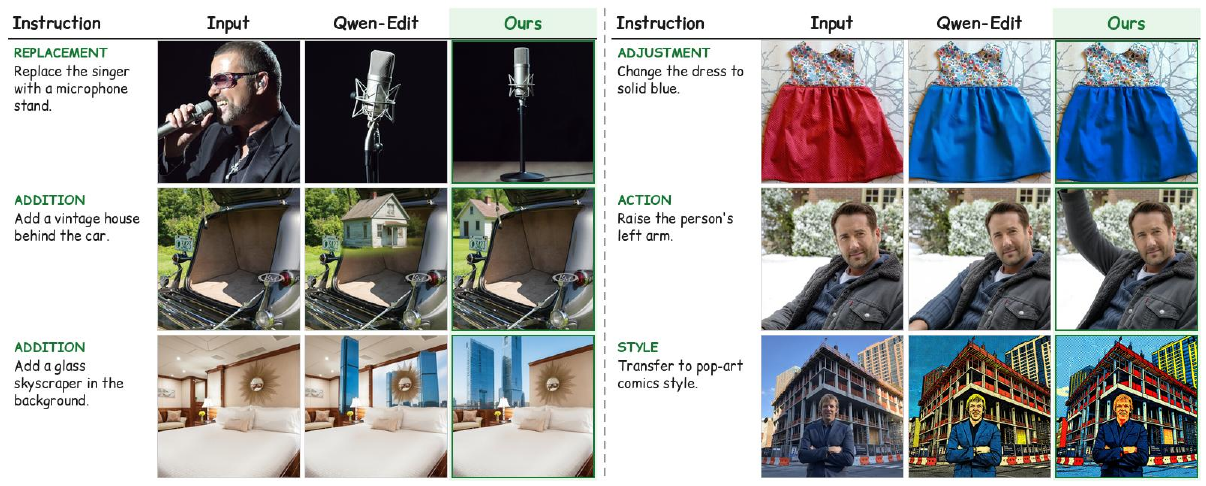}
\caption{\textbf{Qualitative results on edit families beyond object isolation.}
We compare the final \method{} checkpoint with \qwenedit{} on replacement,
addition, adjustment, action, and style-transfer instructions from unused
ImgEdit examples. Self-distillation preserves the baseline's competence on these
families, where the baseline is often already strong, while improving
instruction placement, localization, or visual faithfulness.}
\label{fig:supp_qual_breadth}
\end{figure}

Figures~\ref{fig:supp_qual_high_gain} and~\ref{fig:supp_qual_breadth} present
two complementary qualitative views. The first group contains high-headroom
cases where the matched \qwenedit{} baseline fails to realize the requested
operation and \method{} completes the edit. The second group covers families
where the baseline is often already strong, emphasizing preservation and
cross-family robustness.

%% file: figures/charts/diagnostic_evidence.tex
% Native PGFPlots chart; original scientific data are preserved.
\begingroup\chartmainsetup
\begin{tikzpicture}
\begin{axis}[at={(0.11\linewidth,0pt)},anchor=south west,matched main chart,xmin=.5,xmax=4.5,xtick={1,2,3,4},ymin=0,ymax=1100,ytick={0,500,1000},xlabel={Round},ylabel={Cumulative rows},legend columns=3,legend style={at={(.5,1.03)},anchor=south,font=\chartsmall}]
\addplot[ybar stacked,mark=none,area legend,bar width=\chartmainstackbarwidth,fill=chartGreen,draw=white] coordinates {(1.000000000,76.000000000) (2.000000000,141.000000000) (3.000000000,220.000000000) (4.000000000,302.000000000)};
\addlegendentry{Accepted}
\addplot[ybar stacked,mark=none,area legend,bar width=\chartmainstackbarwidth,fill=chartPurple,draw=white] coordinates {(1.000000000,128.000000000) (2.000000000,272.000000000) (3.000000000,409.000000000) (4.000000000,546.000000000)};
\addlegendentry{Rejected}
\addplot[ybar stacked,mark=none,area legend,bar width=\chartmainstackbarwidth,fill=chartTeal,draw=white] coordinates {(1.000000000,31.000000000) (2.000000000,62.000000000) (3.000000000,94.000000000) (4.000000000,127.000000000)};
\addlegendentry{Replay}
\node[above,font=\chartemph] at (axis cs:4,975) {975};
\node[text=white,font=\chartsmall\bfseries] at (axis cs:4,151) {302};
\node[text=white,font=\chartsmall\bfseries] at (axis cs:4,575) {546};
\node[text=white,font=\chartsmall\bfseries] at (axis cs:4,911.5) {127};
\end{axis}
\node[font=\charttitle,align=center,text width=.40\linewidth] at ({0.11\linewidth+.5\chartmainaxiswidth},115pt) {(a) Cumulative manifest rows};
\begin{axis}[at={(0.585\linewidth,0pt)},anchor=south west,matched main chart,xmin=0,xmax=200,ymin=-.6,ymax=5.6,y dir=reverse,ytick={0,1,2,3,4,5},yticklabels={{Color change},{Color},{Object add},{Tone},{Object remove},{Object replace}},xlabel={Accepted clean targets},xmajorgrids=false,ymajorgrids=false,xtick={0,100,200},yticklabel style={font=\chartsmall}]
\addplot[xbar,mark=none,area legend,bar shift=0pt,bar width=9.5pt,fill=chartGreen,draw=chartGreen,forget plot] coordinates {(158.000000000,0.000000000)};
\node[text=white,font=\chartemph] at (axis cs:79,0) {158};
\addplot[xbar,mark=none,area legend,bar shift=0pt,bar width=9.5pt,fill=chartGreen,draw=chartGreen,forget plot] coordinates {(45.000000000,1.000000000)};
\node[right,font=\chartfont] at (axis cs:45,1) {45};
\addplot[xbar,mark=none,area legend,bar shift=0pt,bar width=9.5pt,fill=chartGreen,draw=chartGreen,forget plot] coordinates {(35.000000000,2.000000000)};
\node[right,font=\chartfont] at (axis cs:35,2) {35};
\addplot[xbar,mark=none,area legend,bar shift=0pt,bar width=9.5pt,fill=chartGreen,draw=chartGreen,forget plot] coordinates {(25.000000000,3.000000000)};
\node[right,font=\chartfont] at (axis cs:25,3) {25};
\addplot[xbar,mark=none,area legend,bar shift=0pt,bar width=9.5pt,fill=chartRed,draw=chartRed,forget plot] coordinates {(3.000000000,4.000000000)};
\node[right,font=\chartemph,text=chartRed] at (axis cs:3,4) {3};
\addplot[xbar,mark=none,area legend,bar shift=0pt,bar width=9.5pt,fill=chartRed,draw=chartRed,forget plot] coordinates {(2.000000000,5.000000000)};
\node[right,font=\chartemph,text=chartRed] at (axis cs:2,5) {2};
\draw[chartInk] (axis cs:0,-.5) -- (axis cs:0,5.5);
% Rejection percentages form a separate annotation column beyond the count axis.
\node[font=\chartsmall] at (axis cs:250,0) {64\%};
\node[font=\chartsmall] at (axis cs:250,1) {52\%};
\node[font=\chartsmall] at (axis cs:250,2) {69\%};
\node[font=\chartsmall] at (axis cs:250,3) {44\%};
\node[font=\chartemph,text=chartRed] at (axis cs:250,4) {92\%};
\node[font=\chartemph,text=chartRed] at (axis cs:250,5) {96\%};
\node[font=\chartsmall] at (axis cs:250,-.95) {Rejected};
\end{axis}
\node[font=\charttitle,align=center,text width=.40\linewidth] at ({0.585\linewidth+.5\chartmainaxiswidth},115pt) {(b) Clean-target bottleneck};
\end{tikzpicture}
\endgroup

%% file: tables/reward_gate_ablation.tex
\begin{table}[!tb]
\centering
\caption{\textbf{Every gate is needed to keep invalid candidates out of training.}
We ablate verifier-bank components on the fixed $210$-pair probe. Acceptance
columns report the final gate rate on valid edits and three negative controls;
AUC columns measure ungated ranking against each negative class. The full
verifier bank accepts $50.5\%$ of valid edits with no false accepts, whereas
removing the editor-side VLM check admits $4.8\%$ of corrupted edits. We
therefore prefer zero false acceptance over higher valid-edit recall, since
every accepted candidate becomes a training target. CEPR only uses the base
Contrastive Edit-Preservation Reward without the rubric checks.}
\label{tab:reward_gate_ablation}

\small
\setlength{\tabcolsep}{4pt}
\begin{tabular}{@{}lccccccc@{}}
\toprule
& \multicolumn{4}{c}{\textbf{Accepted as feasible (\%)}} &
\multicolumn{3}{c}{\textbf{Reward AUC vs.\ valid edit}} \\
\cmidrule(lr){2-5}\cmidrule(l){6-8}
\textbf{Reward variant} & \textbf{Valid} & \textbf{No-op} & \textbf{Corrupt} &
\textbf{Wrong} & \textbf{No-op} & \textbf{Corrupt} & \textbf{Wrong} \\
\midrule
\oursrow \method{} (full verifier bank) & $50.5$ & $0.0$ & $0.0$ & $0.0$ & $0.461$ & $0.916$ & $0.957$ \\
\midrule
CEPR only & $13.3$ & $0.0$ & $0.0$ & $0.0$ & $0.317$ & $0.998$ & $1.000$ \\
w/o forbidden gate & $45.7$ & $0.0$ & $0.0$ & $0.0$ & $0.461$ & $0.916$ & $0.957$ \\
w/o region gate & $61.9$ & $0.0$ & $0.0$ & $0.0$ & $0.472$ & $0.920$ & $0.958$ \\
w/o object detector & $68.6$ & $0.0$ & $0.0$ & $0.0$ & $0.503$ & $0.967$ & $0.991$ \\
w/o editor-side VLM check & $40.0$ & $0.0$ & $4.8$ & $0.0$ & $0.461$ & $0.916$ & $0.957$ \\
relaxed preservation gate & $56.2$ & $0.0$ & $0.0$ & $0.0$ & $0.495$ & $0.910$ & $0.955$ \\
\bottomrule
\end{tabular}

\end{table}

%% file: tables/self_distillation_type_ablation.tex
\begin{table}[!tb]
\centering
\caption{\textbf{Broader reward coverage does not increase the aggregate gain.}
We compare relative ImgEdit gains over the \qwenedit{} base after distilling
either the aligned object-isolation reward or a broader four-type verifier-bank
reward. Both obtain a large extraction gain ($+24.9\%$ and $+25.8\%$), but the
broader reward redistributes secondary improvements, lowering the composition
gain from $+8.4\%$ to $+3.2\%$ and the overall gain slightly from $+5.5\%$
to $+5.3\%$. Covering more edit types is thus not more effective than
focusing on a verifiable one.}
\label{tab:self_distillation_type_ablation}
\small
\setlength{\tabcolsep}{3.6pt}
\begin{tabular}{@{}lcccccccccc@{}}
\toprule
\textbf{Model} & \textbf{Add} & \textbf{Remove} & \textbf{Replace} &
\textbf{Adjust} & \textbf{Action} & \textbf{Style} & \textbf{Bg.} &
\textbf{Extract} & \textbf{Compose} & \textbf{Overall} \\
\midrule
Qwen-Image-Edit-2509 & 4.51 & 4.36 & 4.72 & 4.40 & 4.69 & 4.70 & 4.40 & 3.41 & 4.05 & 4.36 \\
\midrule
\tgroup{11}{$\Delta$ Improvement over base}
\oursrow Object-isolation reward & \uprel{3.1} & \uprel{3.2} & \uprel{3.0} & \uprel{3.2} & \uprel{2.8} & \uprel{2.8} & \uprel{4.3} & \uprel{24.9} & \uprel{8.4} & \textbf{\uprel{5.5}} \\
Four-type verifier bank & \uprel{3.3} & \uprel{3.0} & \uprel{3.0} & \uprel{2.7} & \uprel{3.2} & \uprel{2.8} & \uprel{3.0} & \uprel{25.8} & \uprel{3.2} & \uprel{5.3} \\
\bottomrule
\end{tabular}
\end{table}